\documentclass[10pt,twocolumn,letterpaper]{article}

\usepackage{wacv}
\usepackage{times}
\usepackage{epsfig}
\usepackage{graphicx}
\usepackage{amsmath}
\usepackage{amssymb}
\usepackage{booktabs}

\usepackage{amsmath}
\usepackage{amssymb}
\usepackage{arydshln}
\usepackage{booktabs}
\usepackage{xcolor}
\usepackage{soul}
\usepackage{array,multirow,graphicx}

\usepackage{calrsfs}
\DeclareMathAlphabet{\pazocal}{OMS}{zplm}{m}{n}

\colorlet{soulred}{red!20}
\colorlet{soulgreen}{green!20}
\colorlet{soulblue}{blue!20}
\colorlet{soulorange}{orange!20}

\usepackage[pagebackref=true,breaklinks=true,letterpaper=true,colorlinks,bookmarks=false]{hyperref}

\def\wacvPaperID{1660} 

\wacvfinalcopy 

\ifwacvfinal
\usepackage[breaklinks=true,bookmarks=false]{hyperref}
\else
\usepackage[pagebackref=true,breaklinks=true,colorlinks,bookmarks=false]{hyperref}
\fi

\begin{document}

\title{Designing a Hybrid Neural System to Learn Real-world \\ Crack Segmentation from Fractal-based Simulation}

\author{
Achref Jaziri\\
Goethe University\\
Frankfurt am Main, Germany \\
\texttt{Jaziri@em.uni-frankfurt.de} \\
\and
Martin Mundt\\
TU Darmstadt and  hessian.AI\\
Darmstadt, Germany \\
\texttt{martin.mundt@tu-darmstadt.de} \\
\and
Andres Fernandez Rodriguez\\
University of Tübingen\\
Tübingen, Germany \\
\texttt{a.fernandez@uni-tuebingen.de} \\
\and 
Visvanathan Ramesh\\
Goethe University and  hessian.AI\\
Frankfurt am Main, Germany \\
\texttt{vramesh@em.uni-frankfurt.de} \\
}

\maketitle
\thispagestyle{empty}

\begin{abstract}
   Identification of cracks is essential to assess the structural integrity of concrete infrastructure. However, robust crack segmentation remains a challenging task for computer vision systems due to the diverse appearance of concrete surfaces, variable lighting and weather conditions,  and the overlapping of different defects. In particular recent data-driven methods struggle with the limited availability of data, the fine-grained and time-consuming nature of crack annotation, and face subsequent difficulty in generalizing to out-of-distribution samples. 
In this work, we move past these challenges in a two-fold way. We introduce a high-fidelity crack graphics simulator based on fractals and a corresponding fully-annotated crack dataset. We then complement the latter with a  system that learns generalizable representations from simulation, by leveraging both a pointwise mutual information estimate along with adaptive instance normalization as inductive biases. Finally, we empirically highlight how different design choices are symbiotic in bridging the simulation to real gap, and ultimately demonstrate that our introduced system can effectively handle real-world crack segmentation.
\end{abstract}

\section{Introduction}
The process of structural monitoring and assessment of civil infrastructure is an important task to ensure safety and usability. Executed primarily by humans, the inspection process is time consuming and labor-intensive, as it needs to be carried out at the target location, 
is potentially dangerous and can lead to down-times in the infrastructure use. To alleviate these challenges, the deployment of robots with integrated computer vision systems is emerging as an exciting, safe and low cost addition to traditional inspection methods\cite{koch2015review,verma2013review}.

In general, such a computer vision system should be robust and invariant to a variety of nuisance variables such as illumination, object scale or pose. Early works achieved these desiderata by stacking and combining quasi-invariant transformations specified by a domain expert to guarantee that the output remains unchanged for a range of transformations that are irrelevant to the application domain \cite{binford1993quasi,chin1986model}. In contrast, modern data-driven systems may learn these transformations by relying on large amounts of labeled data. In recent years, deep learning techniques in conjunction with labelled datasets were introduced for structural inspection tasks like crack identification \cite{konig2022s, choi2019sddnet,kang2020hybrid}. 

However, gathering appropriate real-world data for training is tremendously challenging. Data acquisition is particularly tough in the case of cracks on concrete bridges, where defects tend to be located in difficult to capture areas and overlap with other defects like spalling, exposed metal bars etc. Moreover, data labeling is not only excessively time consuming, it is prone to errors due to the fine-grained nature of cracks and requires highly specialized experts (who may not end up agreeing) to provide precise ground-truth \cite{al2020creating}. Previous works have thus proposed datasets addressing the crack identification challenge from a multi-target classification perspective \cite{mundt2019meta}, but similar real-world efforts are still needed for semantic segmentation in diverse contexts.

Faced with a lack of appropriately annotated and diverse data, one may resort to physics based rendering, which has enabled the creation of photorealistic synthetic data for training and testing computer vision models \cite{man2022review,de2021next}. Compared to data-driven based generative models, the approach promises full control over the scenes and automatically generated ground truth maps. Alas, there is a typical statistical mismatch between simulated and real images due to modeling assumptions and computational approximations. Therefore, purely data-driven neural networks trained with synthetic images tend to suffer from performance degradation when applied to real images.

In this work, we seek to overcome existing limitations by combining the strengths of context specific modeling and data-driven designs (hybrid system), enabling us to fully leverage physics based rendering as an underlying source of data. For this purpose, we first introduce a fractal-based concrete crack simulation pipeline and investigate the use of synthetic images for learning in the context of crack segmentation.  We also propose a  model that can achieve better performance by leveraging image-based pointwise mutual information as well as style transfer techniques for better generalization. To empirically examine the latter, we annotate a subset of a prominent real-world concrete defect classification dataset \cite{mundt2019meta} and thoroughly experimentally corroborate our proposed design choices in additional settings. In summary, our contributions are as follows:


\begin{itemize}
   
    \item We propose the ``Cracktal'' high-fidelity simulator for cracks on concrete surfaces to generate pixel wise annotated data with depth and surface normal maps \footnote{We will make the simulator and rendered datasets available on Zenodo.}.
    \item We present an approach to close the gap between performance on simulated and real data through \textbf{C}onsistency enforced training between \textbf{A}daptive instance normalization and \textbf{P}ointwise mutual information, CAP-Net for short\footnote{CAP-Net code will be also open-sourced.}. 
    \item We annotate real-world images of concrete bridges with cracks from the popular CODEBRIM dataset \cite{mundt2019meta} to empirically corroborate our approach.
    \item We investigate the performance of different algorithms for closing the Sim2Real gap in the context of crack detection using a variety of validation metrics specifically tailored for single object semantic segmentation. We empirically validate our approach on the annotated CODEBRIM images as well as other publicly available crack segmentation benchmarks.
    
\end{itemize}

\section{Related Work}
In this section, we summarize related work for crack detection, the use of synthetic data for training neural networks and approaches to reduce the Sim2Real gap.

\subsection{Crack Identification}
Traditional works on crack recognition focus on using image processing algorithms like edge and boundary detection techniques \cite{abdel2003analysis}, morphological operation based methods \cite{yamaguchi2008image}, principle component analysis \cite{abdel2006pca}, or automatic clustering for segmentation based on Canny and K-Means \cite{lattanzi2014robust}. The work of Koch et. al \cite{koch2015review} presents an exhaustive literature review on the common practices of assessing the state of concrete infrastructure and crack detection. 

Recent works leverage data-driven approaches for crack identification using classification or semantic segmentation neural architectures \cite{cheng2018pixel,choi2019sddnet,dais2021automatic}. The works of Cao et. al \cite{cao2020review} and K{\"o}nig et. al \cite{konig2022s} provide a review of current data-driven crack detection approaches. 
However, one of the main limitations of current approaches is that the training data are mostly composed of simple and small datasets with uniform asphalt or concrete backgrounds \cite{zhang2016road,eisenbach2017how,shi2016automatic,amhaz2016automatic,zou2012cracktree}, 
which hinders effective generalization of data-driven approaches in the context of precise semantic crack segmentation. In our work, we overcome the data hurdles by proposing a high-fidelity data simulator, which we can fully leverage by proposing a model that incorporates necessary inductive biases while enabling effective learning.

\subsection{Simulating Data and the Sim2Real Gap}

In recent years, data-driven generative models have gained considerable popularity. Despite that, simulators based on physics-based rendering engines have maintained their significance. This is largely attributed to their ability to effortlessly produce pixel-accurate labels, thus reducing the burden of manual annotation. Furthermore, these simulators offer a unique advantage in generating data with controlled priors, enabling the generation of diverse datasets tailored to specific scenarios and applications. Proposed simulators in the literature include GTA5 \cite{richter2016playing}, SYNTHIA \cite{ros2016synthia} and endless runner for continual learning \cite{hess2021procedural}. 

Whereas some works show promising results for the use of synthetic data in detection tasks \cite{varol2017learning,mu2020learning,wang2019learning}, models trained with synthetic data are well-known to face difficulties in generalizing to real data, due to the statistical gap between synthetic images and real images \cite{vazquez2013virtual,vaudrey2008differences,xu2014domain}. Apart from improving the graphics rendering pipeline itself, this Sim2Real gap is typically reduced by seeking out domain adaptation (DA) or domain generalization (DG) techniques. 

DA approaches focus on adapting the statistics of the synthetic data to that of the target domain, for instance by adversarially tuning the parameters of the generative models based on the statistics of the real data for better generalization \cite{veeravasarapu2017adversarially,alhaija2018geometric}. Others \cite{kim2020learning} make use of style transfer methods to adapt the training data. The main limitation of DA techniques remains that they require access to samples in the target domain in order to adapt the model. We refer to \cite{zhao2020review,wang2018deep} for comprehensive surveys. 

In contrast, DG approaches seek to improve the robustness of DNNs to arbitrary unseen domains, see Wang et. al \cite{wang2022generalizing} for a detailed review.  Approaches for learning domain-agnostic feature representations can leverage meta-learning \cite{balaji2018metareg,dou2019domain}, adversarial training \cite{li2018domain}, instance normalization \cite{peng2022semantic}, selective whitening \cite{choi2021robustnet}, style transfer or data augmentation \cite{yue2019domain}. 
 
\begin{figure*}[t]
	\centering
	\includegraphics[width=0.225\linewidth]{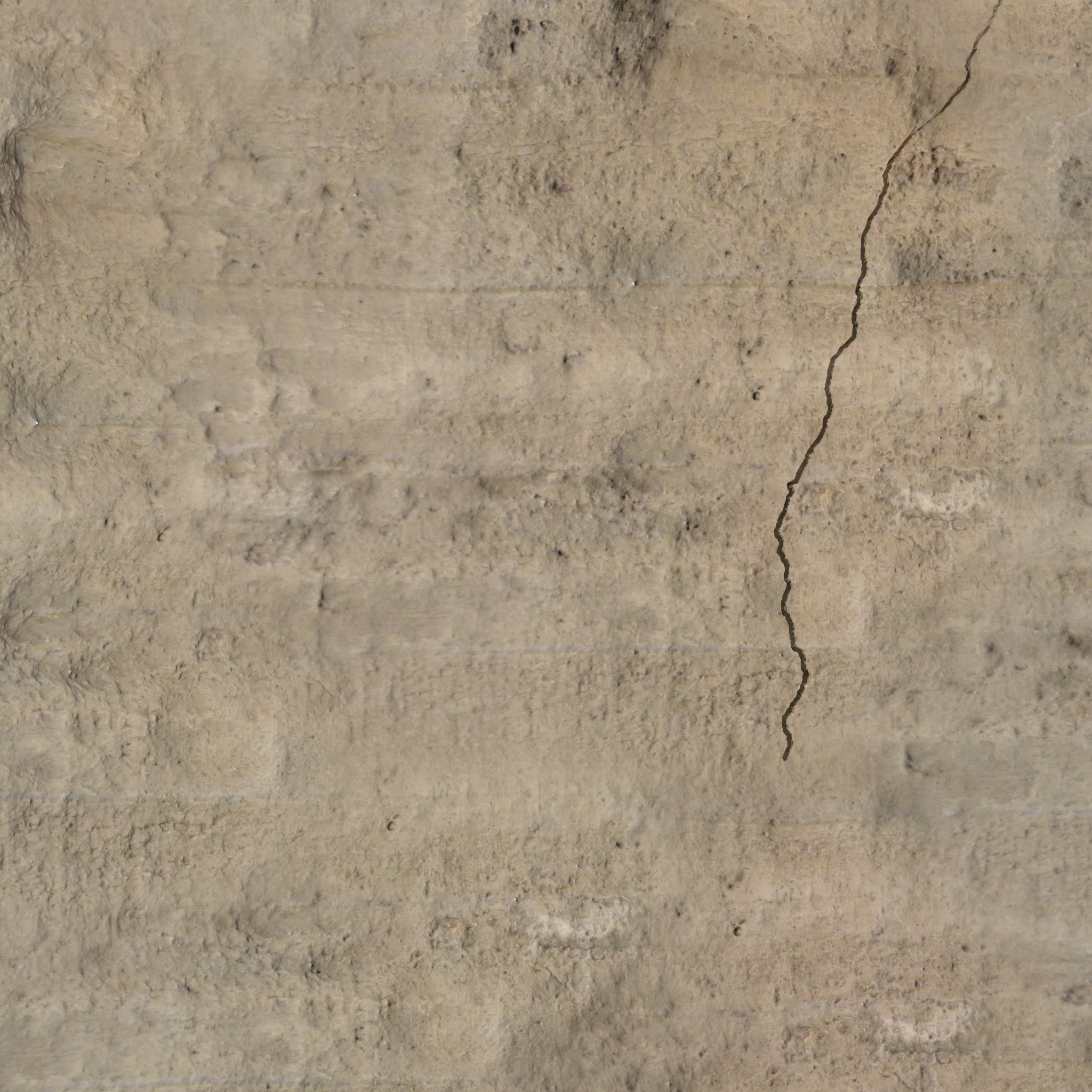} \hspace{0.001cm}
    \includegraphics[width=0.225\linewidth]{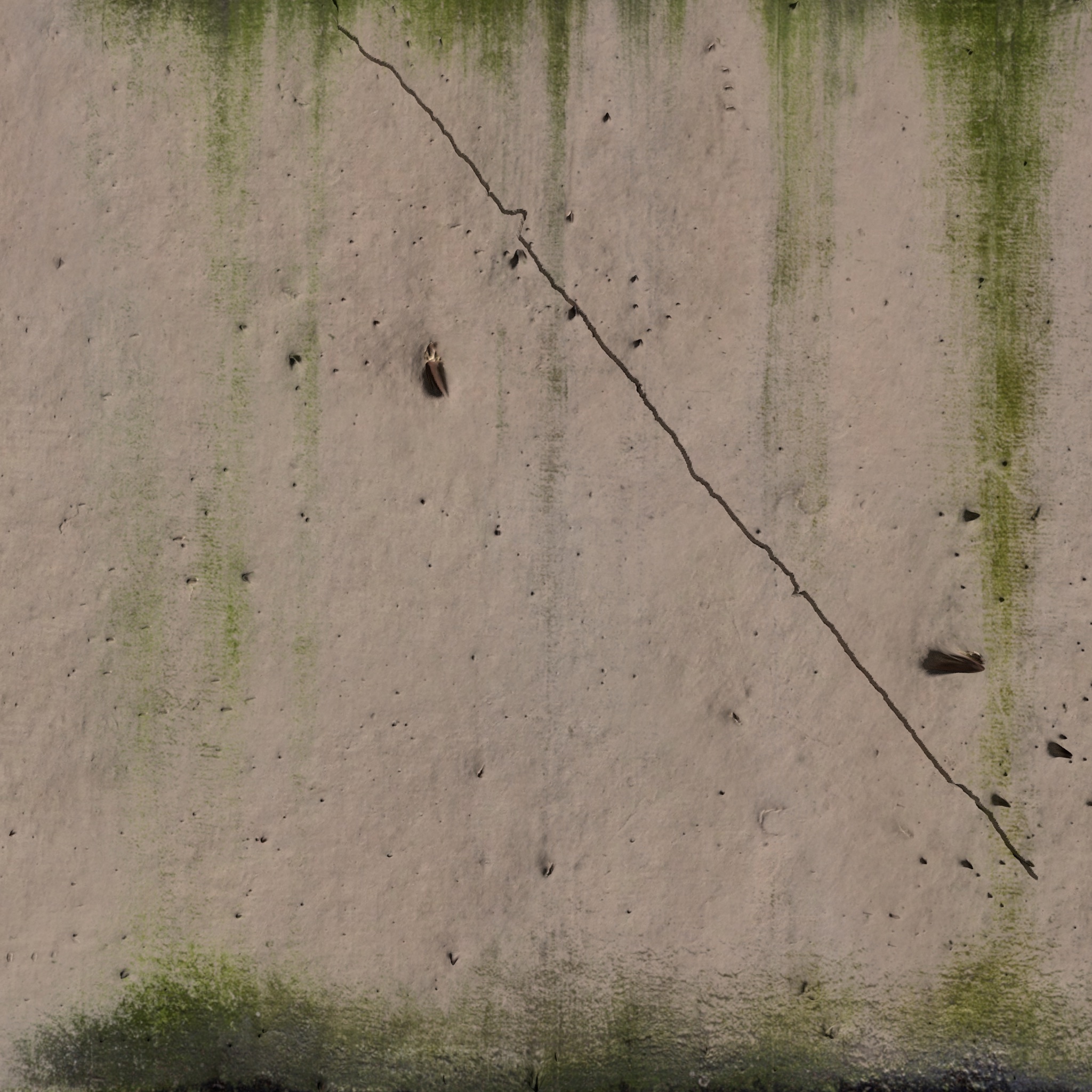}
    \hspace{0.001cm}
    \includegraphics[width=0.225\linewidth]{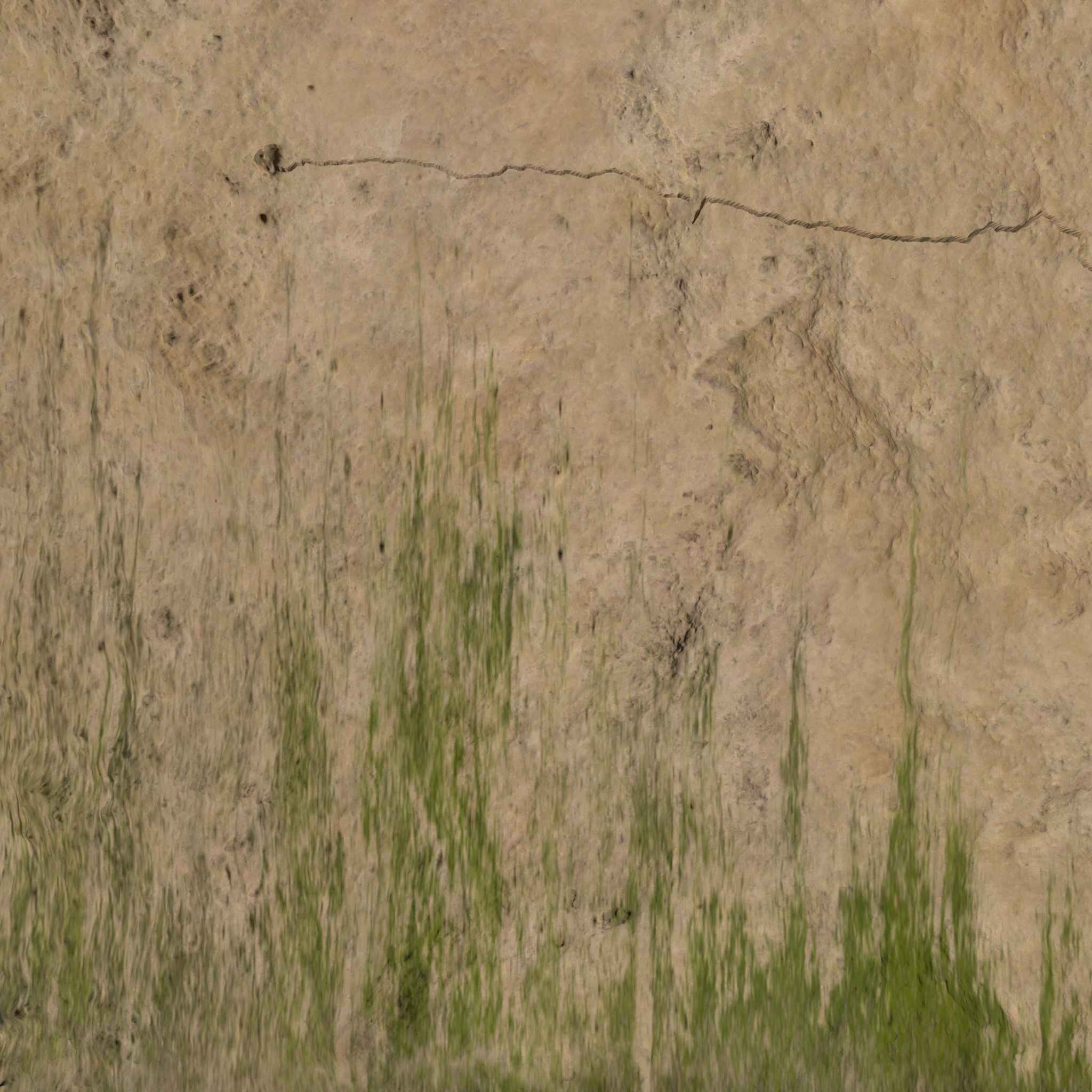}
    \hspace{0.001cm}
    \includegraphics[width=0.225\linewidth]{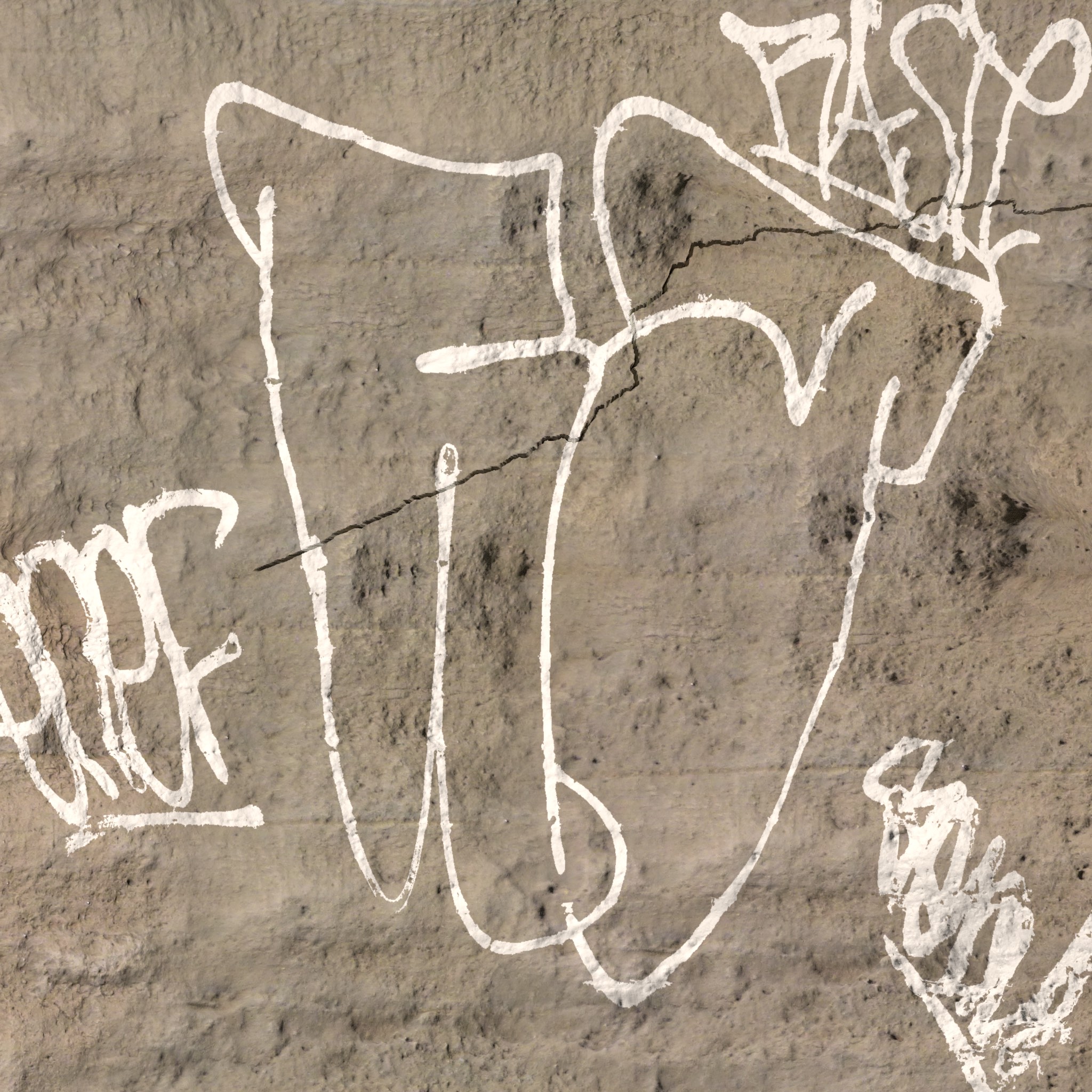}
	\caption{Cracktal examples with texture variety and presence of other perturbations/ anomalies like moss and graffiti. Images were generated at $2048 \times 2048$ resolution and are heavily down-sampled for view in pdf at loss of quality.}
	\label{SimExamples}
\end{figure*}

Other works bias their models to focus on image features that are more realistic and transferable to improve generalization in a given application domain. For instance, features related to the image geometry can be more generalizable in the case of car detection \cite{saleh2018effective}. Since geometry and semantics are naturally connected,   \cite{chen2019learning, jiang2017configurable} propose to mitigate the limitations of synthetic data by leveraging the geometric information in a multi-task learning framework. In our work, we follow the spirit of these works, but note that most application contexts of synthetic data in the literature focus on objects with well defined shapes (e.g cars, buildings etc.). In contrast, we emphasize that cracks, like most defects, have highly irregular shapes. We design an extendable physics-based crack simulator and subsequently leverage specific crack sensitive quasi-invariant models to learn more generalizable representations in our CAP-Net approach to reduce the Sim2Real gap.

\section{Cracktal: A Fractal-based Simulator for Cracked Concrete Surfaces}

In this section, we introduce Cracktal: a physics-based simulator that generates images of cracked concrete surfaces along with their semantic ground truth, depth and surface normal maps. The overall rendering workflow consists of two main steps: scene and crack generation.
A set of albedo, roughness, normal and height maps are used to set the scene based on physics based rendering rules. A random crack is then generated using our fractal generator model, detailed below, and added to the scene's material. The full scene is then rendered and corresponding ground truth maps are generated. Figure \ref{SimExamples} illustrates examples of the synthetic images generated with a $2048 \times 2048$ resolution. 

\subsection{Physics-based Scene Generation}
In the scene generation process, non relevant backgrounds (e.g sky, out of focus buildings etc.) are excluded, assuming an up-close camera. The base components of physics based rendering (PBR) workflows: albedo, normal, roughness, and height maps are applied to a plane mesh grid to generate a realistic looking concrete surface, defining its color, surface and subsurface scattering, and geometrical displacement respectively. The required PBR metallicity map is included but remains uniformly zero, as concrete is a dielectric. An optional ambient occlusion map can be included to introduce surface markings, e.g. graffiti. The textures used in this work were created from real concrete images from the CODEBRIM training dataset \cite{mundt2019meta} and decomposed manually by the Substance B2M software.

The environment is illuminated utilizing a simulated natural sunlight source. The black body radiator  possesses two key attributes: luminous intensity ($L_I$) and color temperature ($L_T$). Luminous intensity determines the amount of energy that the light source emits into the scene, whereas color temperature defines the chromaticity of the illuminant. In the datasets of our later study, a color temperature of $L_T= 5800$ Kelvin and intensity $L_I=3.3$ were chosen. The rotation of the light source is parameterized by its Euler $ \alpha$, $\beta$ $ \gamma$ angles as in common conventions. $ \alpha$ and $ \gamma$ are fixed angles with values $ \frac{\pi}{3}$ and $ 0 $. By varying the $ \beta$ angle, we simulate the change of the hour of day during which the image is captured. The $\beta$ angle is randomly sampled from:
\begin{align}
\beta \sim \mathcal{U}(\frac{-\pi}{6}, \frac{\pi}{6})
\end{align}
\subsection{Fractal-based Crack Generation}
Cracks are highly irregular, but like many other patterns found in nature can be represented as fractals. 
In order to generate a crack pattern, we  draw inspiration from a decades old model presented by \cite{leblanc1991analysis} as a baseline.  The authors suggest the use of a stochastic version of the Koch ''snowflake`` fractal in the generation of pavement distress features, e.g cracks on a road surface. A conventional Koch ''snowflake`` fractal can be generated through the iterative splitting of each straight line into three equal length segments. The middle segment is then replaced by two segments of equal length to form an equilateral triangle. These steps are repeated for each straight line to create a regular fractal until a desired subdivision depth is reached. 

\begin{figure}[t]
	\centering
	\includegraphics[width=\linewidth]{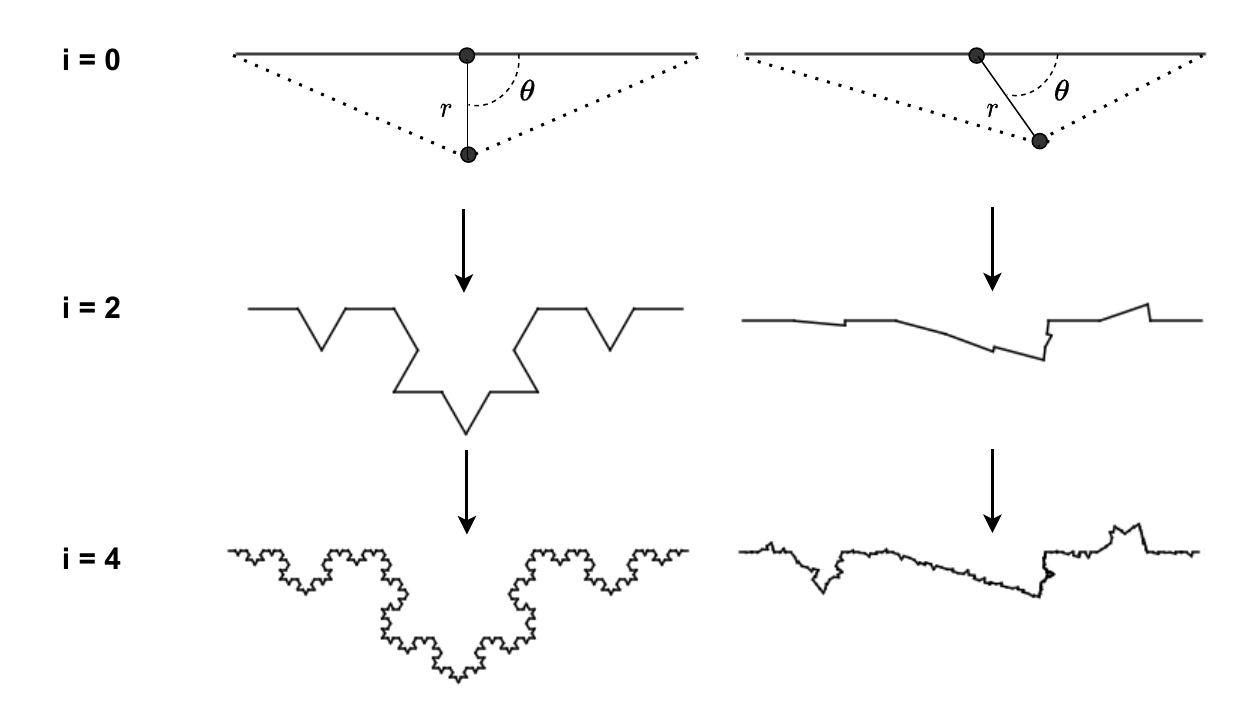}
 
	\caption{Example Koch fractal (left) and stochastic version for cracks (right). While the magnitude $r$ and the angle $\theta$ of the displacement are typically fixed, they are randomized for each displacement iteration (i) when generating the irregular crack shape. }
	\label{Fractal}
\end{figure}

By modifying the displacement parameters at each step, it is feasible to generate non-uniform fractals that resemble cracks. Rather than dividing each line into identical segments to form an equilateral triangle, the position of the third point, which is determined by both the magnitude $r$ and angle $\theta$, is altered in each step. In our simulation, the angle is sampled from a Gaussian normal distribution with a mean of $\mu=0$ and a standard deviation of $\sigma=30$ degrees. The probability density of displacement magnitude r is given by $P(r)= \frac{2r}{p^2}$ where p is a hyper-parameter. For intuition, Figure \ref{Fractal} illustrates these steps for a ''snowflake`` and the stochastic version for crack generation. 

Finally, before adding the crack to the rendered scene, the generated crack is randomly translated and rotated across the scene,  and a Gaussian blur is applied in order to introduce width to the crack.

\subsection{Annotation of Real World Data}

In conjunction with simulated data, we require real world images to validate systems trained with synthetic images. Ideally, the chosen real world image should offer additional challenges that make generalization less straightforward. For these reasons, we semantically annotated images provided in the CODEBRIM dataset \cite{mundt2019meta} on a pixel basis.  

We have chosen this dataset as its development has been motivated by the need for a concrete visual inspection dataset that contains other overlapping defects and features various levels of deterioration, defect severity and surface appearance. Previous works \cite{zhang2016road, eisenbach2017how, shi2016automatic, amhaz2016automatic, zou2012cracktree}, focused on data where cracks are the only visible defect, and they are usually centered in the image, making them unrealistically easy to segment. Most of them also show pavement cracks, which may differ in appearance from concrete cracks. In many CODEBRIM images, other defects like exposed reinforcement bars, spallation, corrosion and calcium leaching are present. In particular the latter share visual similarities with cracks, which makes the prediction more challenging. 

We selected image patches containing visible cracks and annotated them using GIMP. 
In this way, Multiple annotators semantically annotated images of $1500 \times 844 $ resolution, each containing at least one crack. We consolidated consistent annotations into a set of 420 examples for our real-world test set.

\section{CAP-Net: A Hybrid Neural Approach for Crack Segmentation}
\begin{figure*}[h!]
	\centering
    \includegraphics[width=\linewidth]{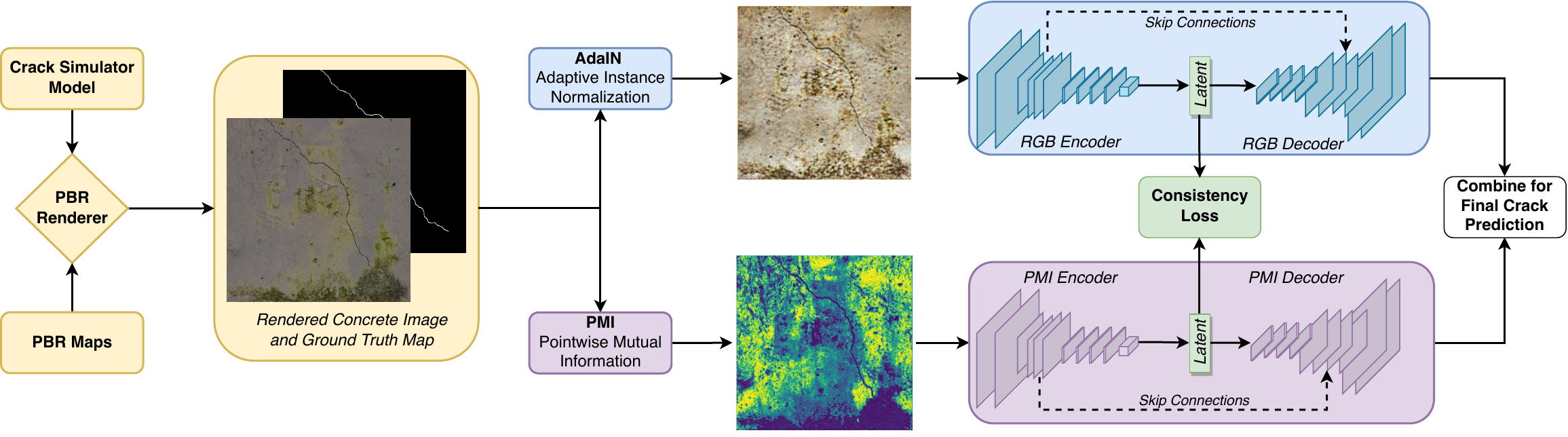}%
	\caption{Schematic of the Cracktal to CAP-Net training pipeline. Based on simulated images and their automatic annotation (yellow shading), consistency enforced training (green) is performed between two networks extended with adaptive instance normalization (AdaIN, blue) and pointwise mutual information (PMI, purple) respectively. For final inference, the AdaIN style-transfer module is dropped.}
	\label{mainPipeline}
\end{figure*}

To fully leverage our simulator and  bridge the Sim2Real gap, we introduce  \textbf{CAP}-Net, a hybrid neural model based on \textbf{C}onsistency enforced training between \textbf{A}daptive instance normalization and \textbf{P}ointwise mutual information. It is composed of two parallel network branches, each one based on a U-Net architecture \cite{ronneberger2015u}. During training, we input the RGB image stylized by the AdaIN module. The second network is equipped with a PMI module to extract representations that are projected into a quasi-invariant feature space that helps with the domain transfer. Both networks are connected with a consistency loss to enforce common representations across the different domains. We train our pipeline end to end with the help of the synthetic images and their ground truths generated by Cracktal, as depicted in Figure \ref{mainPipeline}.

\subsection{Pointwise Mutual Information}
Cracks can be viewed as anomalies in a textured surface. 
Pointwise mutual information (PMI), computed in a local neighborhood, is a measure of deviation of the gray-level co-occurrence statistics in the neighborhood relative to marginal statistics of gray-levels globally.  Thus, the PMI measure flags boundaries between dominating texture patterns in the image. The resulting output is an indicator of texture anomalies and  directly relates to hypothesized cracks and other boundary structures.


Drawing inspiration from prior work \cite{isola2014crisp}, natural objects produce probability density functions that are well clustered. These clusters can be discovered in an unsupervised manner, and fitted by kernel density estimation (KDE). The obtained density functions can then further be leveraged to distinguish between common pixel pairs (belonging to the background texture) from less common pairs (belonging to anomalies in the texture or edges). To compute the PMI scores between two pixels, we first need to estimate the joint distribution and marginal distributions for image pixels.  
For the marginal distribution  $\mathit{P(A)}$ , we sample pixels randomly from the image to perform the KDE. To estimate the joint distribution $\mathit{P(x_i,x_j)}$, we sample pairs of pixels at various distances and perform KDE. 

For a pixel pair $(x_i,x_j)$ the PMI score is computed as: 
\begin{align}
    \mathit{PMI(x_i,x_j)} = \mathit{log} \frac{P(x_i,x_j)^{\tau}}{P(x_i) \cdot P(x_j)}
\end{align}

The  parameter $\tau$ boosts the scores of common pairs and addresses the bias of PMI towards low-frequency events (i.e when the marginal distributions are small). When $\tau = 1$, $ \mathit{PMI(x_i,x_j)}$ specifically  compares the likelihood of observing the pixel $x_i$ near $x_j$ to the overall probability of observing $x_i$ and $x_j$ in the image. The final affinity score for each pixel is computed using the PMI score with the neighboring pixels.  We define the set of neighbouring pixels of $x_i$ as $N_i$. 
The PMI scores between a pixel and its neighbors are exponentiated and summed to estimate an affinity score for each pixel in the original image, indicating if this pixel belongs to the dominant background texture or is an anomaly:
\begin{align}
\mathit{ Affinity(x_i)} = \sum_{x_j \in N_i} e^{PMI(x_i,x_j)}
\end{align}
The scores are then passed to a neural network for crack prediction. Note that the exponential is important to obtain more stable affinity scores, which helps with learning.

\subsection{Style Transfer}
In addition to the use PMI as an inductive bias, we further reduce the Sim2Real gap from the data-driven angle by  performing style transfer operations based on the adaptive instance normalization (AdaIn) \cite{huang2017arbitrary}, which aligns the mean and variance of the content features with those of the style features.  
The content features are obtained by encoding an image generated by Cracktal simulator using a VGG network pretrained on ImageNet \cite{deng2009imagenet}. Similarly, the style features are encoded from a texture image. The AdaIN layer is used to perform style transfer in the feature space and aligning the features of the content and style images. A decoder is learned to invert the AdaIN output to the image spaces.

We sample texture from the describable textures dataset \cite{cimpoi14describing} to perform the style transfer on Cracktal images. This way, we can augment the synthetic training data  and increase the texture variety while at the same time keeping the semantic content of the original image and more specifically the crack intact. We note that style transfer is only performed during training with probability of $0.5$ and is completely dropped during testing.

\subsection{Consistency Loss} 
Finally, to get the best of both worlds, we add a consistency loss between the network trained with RGB images and the network trained with PMI based affinity scores. We postulate that ensuring consistency of the latent space representations across projected subspaces of the outputs of two networks will lead to robust features that will enable better transfer to real data.
For training image $X_i$, The consistency loss is imposed as  follows:

\begin{align}
\mathit{\pazocal{L}_{CL}(X_i)} = (f_1(\mathit{enc_{rgb}(X_i)})-f_2(\mathit{enc_{pmi}}(X_i)))^2
\end{align}

where $\mathit{enc_{rgb}}$ and $\mathit{enc_{pmi}}$ are the encoding functions for the RGB and PMI networks respectively. The obtained latent encodings are then passed to projection heads ($f_1$ and $f_2$) before contrasting them.

\section{Experiments}
Our empirical investigation follows four key questions: 

\noindent \textbf{(Q1) Are Cracktal assumptions plausible?} To corroborate the plausibility of our modelling assumptions and the utility of synthetic data for crack detection, we contrast the performance of a U-Net trained with real world publicly available data with a U-Net trained with our synthetic data. 

\noindent  \textbf{(Q2) Do simulated auxiliary tasks improve generalization?} In the spirit of prior works \cite{chen2019learning, jiang2017configurable}, we further consider how the addition of auxiliary tasks to the baseline U-Net can improve generalization performance in the context of crack segmentation. More specifically, crack patterns have local geometric variations, e.g. surface normal distribution variation, and depth variations relative to the geometry and depth in the surrounding context. Similarly, PMI maps address estimation of an auxiliary task of appearance anomaly extraction.

\noindent  \textbf{(Q3) Does our approach of CAP-Net reduce the Sim2Real gap?}  We empirically corroborate that our proposed method outperforms existing baselines, even when the latter are trained on real-data, effectively demonstrating how our design choices along with a domain specific simulation can lead to more robust crack segmentation models. 

\noindent  \textbf{(Q4) Are all design choices for CAP-Net meaningful?}  We ablate each component of our hybrid CAP-Net to showcase that each proposed element has meaningful impact towards the overall CAP-Net performance. \\


\subsection{Baselines and Additional Evaluation Datasets}

In addition to SegCODEBRIM, we evaluate our models on a collection of the following public datasets: CRACK500 \cite{zhang2016road}, GAPs384 \cite{eisenbach2017how}, CFD \cite{shi2016automatic}, AEL \cite{amhaz2016automatic}, Cracktree200 \cite{zou2012cracktree}. We merge these into 950 images of cracks captured under various conditions. We refer to the experiments using these datasets collectively as the multi-source set. For consistency,
We downsample all images to $256 \times 256$. As intuitive baselines, we consider the following models: A U-Net trained with synthetic data (U-Net), A U-Net trained with collection of multi-source data (U-Net(MultiSet)), A U-Net trained with real and synthetic data (U-Net(Sim+Real)). In addition, we compare to the attention based U-Net variant (Attn-U-Net) \cite{oktay2018attention} and to TransU-Net \cite{chen2021transunet} that combines transformer-based architectures with U-Net. For analysis of the multi-task training in Q2, we further construct Multi-U-Net architectures, based on a single joint encoder and one separate decoder per modality, in the spirit of prior segmentation works outside the crack defect application \cite{chen2019learning, jiang2017configurable}.  



\subsection{Evaluation Metrics}
Evaluating binary semantic segmentation maps with common overlap based scores such as Dice or Intersection over Union (IOU) comes with various limitations. For cracks, connectivity is important but slight over- or under- segmentation of crack pixels can be tolerated, especially knowing that the ground truth maps are usually annotated by humans using different annotation tools with varied settings. For these reasons, we take inspiration from the medical imaging literature and adapt various metrics to obtain more insights into our models \cite{ma2021loss,karimi2019reducing}. \\ 

\noindent \textbf{Hausdorff based Metrics \cite{huttenlocher1993comparing}:}
For two point sets X and Y, the one-sided Hausdorff Distance from X to Y is:
\begin{align}
\mathit{hd(X,Y)} = \max_{x \in X} \min_{y \in Y} dist(x,y)
\end{align}
where $dist$ is a distance measuring function between pixels x and y. The bidirectional Hausdorff Distance is then:

\begin{align}
\mathit{HDF(X,Y)} = \max(hd(X,Y),hd(Y,X))
\end{align}
We use both the euclidean distance and radial basis function (RBF) as a distance measure between pixels. RBF, also known as the squared exponential kernel, is defined as: 

\begin{align}
\mathit{RBF(x,y)} = exp(- \frac{d(x,y)^2}{2l^2})
\end{align}
where d is the euclidean distance between x and y. A main advantage of using RBF as a distance measure is that it decreases gradually the further the prediction is from the ground truth, whereas Dice-scores or IOU decays completely regardless of the distance between the actual prediction and the ground truth. In the case of cracks, this distance could be just very few pixels. \\

\noindent \textbf{clDice:} The authors of \cite{shit2021cldice} introduce a similarity measure centerlineDice (clDice), calculated by comparing the intersection of the prediction and ground truth masks  and their morphological skeleta. Given two binary segmentation maps, ground truth $GT$ and prediction $P$, $S_{GT}$ and $S_{P}$ are the respectively extracted skeletons. Subsequently, the fraction of $S_X$ that lies within $Y$ (Topology Precision), and vice-a-versa (Topology Sensitivity) are:
\begin{align}
    T_{prec}(S_{P},GT)=\frac{\mid S_{P} \cap GT \mid }{S_{P}}
\end{align}

\begin{align}
    T_{sens}(S_{GT},P)=\frac{\mid S_{GT} \cap P \mid }{S_{GT}}
\end{align}
These can then be used to define the clDice score:

\begin{align}
    clD(GT,P)=2 \times \frac{T_{prec}(S_{P},GT) \times T_{sens}(S_{GT},P) }{T_{prec}(S_{P},GT)+ T_{sens}(S_{GT},P)}
\end{align}

\noindent \textbf{\boldmath $F1_{\theta}$:} We also consider a F1 scores with tolerance measure. In the experiments in the main body, we set $\theta=10$.

\subsection{Results and Discussion}
\begin{table*}[t]
  \centering
  \begin{tabular}{@{}llcccccc@{}}
     & Model &$F1 (\uparrow)$ & $F1_{\theta=10} (\uparrow)$  & $clDice (\uparrow)$  & $HDF_{Euc} (\downarrow)$ & $HDF_{RBF} (\downarrow)$ \\ 
    \midrule
   \parbox[t]{2mm}{\multirow{6}{*}{\rotatebox[origin=c]{90}{SegCODEBRIM}}} & U-Net  \cite{ronneberger2015u}     &   { $29.6 \pm 3.1$} &    { $35.8 \pm 1.8$}  & { $38.5 \pm 0.9$}  &  { $40.2 \pm 13.2$} & { $53.6 \pm 7.1$} \\
   & U-Net (MultiSet)       &   { $25.6 \pm 1.7$} &    { $37.2 \pm 0.9$}  & { $28.5 \pm 3.7$}   &  { $53.9 \pm 10.7$} & { $66.5 \pm 1.2$} \\
    & U-Net (Sim+Real)       &   { $33.4 \pm 3.8$} &    { $35.5 \pm 4.4$}  & { $51.3 \pm 4.2$}   &  { \boldmath $23.5 \pm 3.6$} & { \boldmath $42.7 \pm 5.3$} \\
   & Attn-U-Net \cite{oktay2018attention}      &   { $31.2 \pm 1.1$} &    { $37.5 \pm 4.5$}  & { $43.1 \pm 6.2$}  &  { $34.7 \pm 10.6$} & { $51.7 \pm 5.5$} \\
   & TransU-Net \cite{chen2021transunet}       &   { $28.4 \pm 1.8$} &    { $31.3 \pm 2.3$}  & { $43.6 \pm 2.0$}  &  { $25.6 \pm 1.1$} & { $46.8 \pm 2.2$} \\
    
   & Multi-U-Net (D-SN)      &   { $31.9 \pm 1.3$} &    { $36.4 \pm 0.9$}  & { $45.3 \pm 1.2$}  &  { $30.5 \pm 7.1$} & { $46.4 \pm 4.2$} \\

  &  Multi-U-Net (PMI)      &   { $32.7 \pm 0.8$} &    { $37.1 \pm 1.1$}  & { $47.2 \pm 1.7$}  &  { $25.7 \pm 6.4$} & { $43.9 \pm 3.3$} \\

  &  CAP-Net &   {\boldmath $37.3 \pm 1.5$} &    {\boldmath $40.4 \pm 1.8$}  & {\boldmath $53.6 \pm 1.5$} &  {\boldmath $23.3 \pm 1.0$} & {\boldmath $42.6 \pm 1.8$} \\
    
    \bottomrule
  \end{tabular}
  \begin{tabular}{@{}llcccccc@{}}
     \parbox[t]{2mm}{\multirow{6}{*}{\rotatebox[origin=c]{90}{Multi-source Set}}} & U-Net \cite{ronneberger2015u}      &   { $42.6 \pm 0.2$} &    { $44.5 \pm 0.2$}  &   { $71.5 \pm 0.5 $}  &  { $13.7 \pm 1.3$} & { $33.5 \pm 1.0$} \\
    & U-Net (Sim+Real)       &   { $41.4 \pm 3.7$} &    { $48.5 \pm 3.7$}  & { $57.5 \pm 6.9$}   &  { $45.1 \pm 5.4$} & { $49.6\pm 4.1$} \\
   &    Attn-U-Net   \cite{oktay2018attention}    &   { $41.2 \pm 2.2$} &    {$43.2 \pm 2.6$}  &    {$69.4 \pm 1.7$}  &  {$14.9 \pm 1.0 $} & {$35.3 \pm 2.7$} \\
    & TransU-Net \cite{chen2021transunet}      &   { $28.4 \pm 1.8$} &    { $31.3 \pm 2.3$}  & { $43.6 \pm 2.0$}  &  { $22.0 \pm 1.1$} & { $46.9 \pm 2.2$} \\
   &   Multi-U-Net (D-SN)     &   { $42.5 \pm 2.9$} &    { $44.9 \pm 2.5$}  &   { $68.9 \pm 3.8$}  &  { $13.5 \pm 1.8$} & { $33.6 \pm 1.2$} \\

   &   Multi-U-Net (PMI)      &   { $41.0 \pm 2.1$} &    { $43.1 \pm 1.5$}  &   { $66.9 \pm 5.8$}  &  { $14.1 \pm 1.4$} & { $32.7 \pm 1.2$} \\  
  &  CAP-Net &   {\boldmath $44.9 \pm 1.5$} &    {\boldmath $46.9 \pm 1.8$}  & {\boldmath $72.30 \pm 1.5$} &  {\boldmath $13.2 \pm 1.0$} & {\boldmath $31.5 \pm 1.8$} \\
\cline{2-7}
  &    U-Net (MultiSet) $*$       &   {\boldmath  $68.4 \pm 0.3$} &    {\boldmath  $82.8 \pm 0.6$}  &     {\boldmath   $88.4 \pm 1.1 $}  &  { \boldmath $5.7 \pm 1.4$} & { \boldmath  $13.9 \pm 0.8 $} \\
  \end{tabular}
  \smallskip
  \caption{Performance comparison of different models on SegCODEBRIM (top half) and multi-source set (bottom half). The best performing models on each dataset are highlighted in bold, where multiple values are highlighted if they lie within statistical deviations. CAP-Net outperforms \emph{all} baselines on the real-world SegCODEBRIM, despite only being trained on simulated Cracktal data. It even outperforms the MultiSet trained on real multi-source data. Note that we also mark both CAP-Net and U-Net (MultiSet) for the multi-source dataset, as the latter is trained on real-world in-domain data and provides an expectation of what could be achieved, which we mark with a $*$. Apart from this upper-bound, CAP-Net beats all other simulation based baseliness.}
  \label{tableRes1}
\end{table*}


\noindent \textbf{(Q1) The modelling assumptions in Cracktal are plausible:} The top half of table \ref{tableRes1} shows the performance of models when evaluated on SegCODEBRIM. Despite training with real-world data and annotations of the multi-source dataset, U-Net (MultiSet) achieves an F1 score of $25.6 \%$, which is worse than the performance of U-Net trained with synthetic Cracktal data. A similar trend can be observed on all metrics except $F1_{\theta=10}$, where U-Net (MultiSet) outperforms the baseline U-Net only marginally. We hypothesize that the overall worse performance achieved by U-Net (MultiSet) can be explained by the fact that the used training dataset comes from a variety of sources that tend to feature inconsistent annotation styles. More generally, U-Net (MultiSet) achieves a higher number of false positives and detects other anomalies present on concrete surfaces compared to U-Net, as evidenced by clDice and Hausdorff distance scores. These results underscore the significance of accurate labeling, which is guaranteed in simulation. Thus, we find the plausibility of our modelling assumptions in the Cracktal simulator to be well supported. 

\noindent \textbf{(Q2) Auxiliary simulated tasks improve the generalization:} We consider two auxiliary tasks: depth and surface normals prediction and estimation of pointwise mutual information, denoted by the trained Multi-U-Net (D-SN) and Multi-U-Net (PMI) in table \ref{tableRes1} respectively. Both models outperform the baseline U-Net significantly, improving clDice by $7+$ and the Euclidean Hausdorff distance measure by $10+$ on SegCODEBRIM dataset. Similar trends can be observed for the other metrics. Clearly, the depth and surface normal maps predicted by Multi-U-Net (D-SN) provide valuable information about the 3D spatial structure and layout of the scene, thus improving generalization on real data. Similarly, estimating geometry information can be seen as an inductive bias; Multi-U-Net (PMI) learns representation that focus on the anomalies in the images and cracks can be also understood as anomalies. 

 However, both of these models are less robust than our CAP-Net, highlighted also by the fact that the Multi-U-Nets do not significantly outperform the baseline U-Net when evaluated on the  multi-source data (bottom half of Table \ref{tableRes1}). \\

\noindent \textbf{(Q3) CAP-Net's hybrid modelling effectively reduces the Sim2Real gap:} Revisiting Table \ref{tableRes1}, CAP-Net clearly outperforms all approaches on SegCODEBRIM (top half of table). For instance, we observe an improvement of $7 \%$ in F1 and $11$ in Hausdorff distance with RBF kernel compared to U-Net. Similarly, our model performs better than all the baselines on multi-source set (bottom half of table), except U-Net(Multiset), which has been trained with in-distribution training data. Overall, training on the real multi-source data is only beneficial when deploying in a closely related context, whereas the  modelling of CAP-Net in conjunction with the Cracktal simulator provides a robust solution for widely applicable crack segmentation by adapting from purely synthetic data. \\


\begin{figure*}[htbp]
	\centering
	\includegraphics[width=0.78\linewidth]{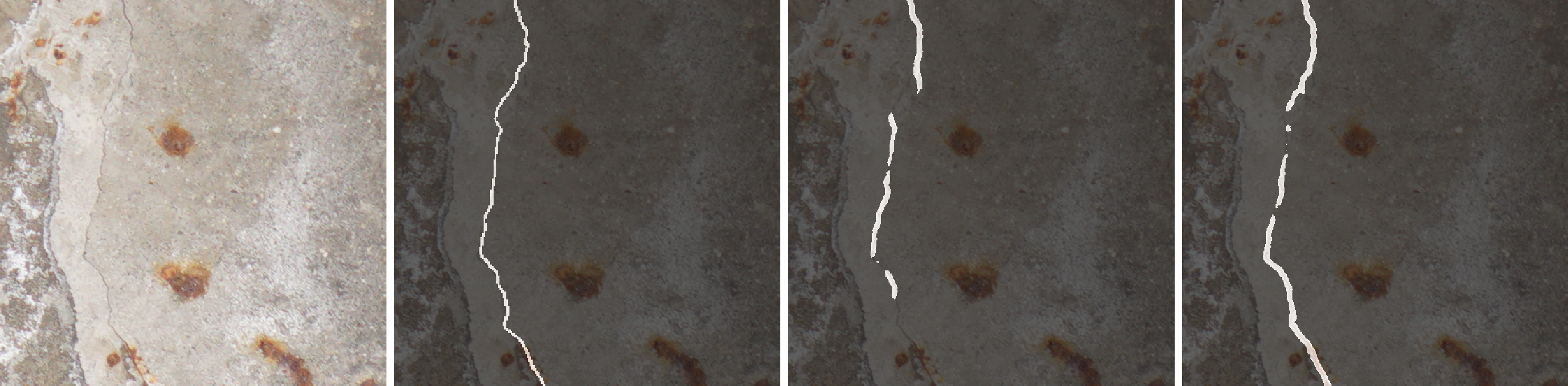}
 \\
 \includegraphics[width=0.78\linewidth]{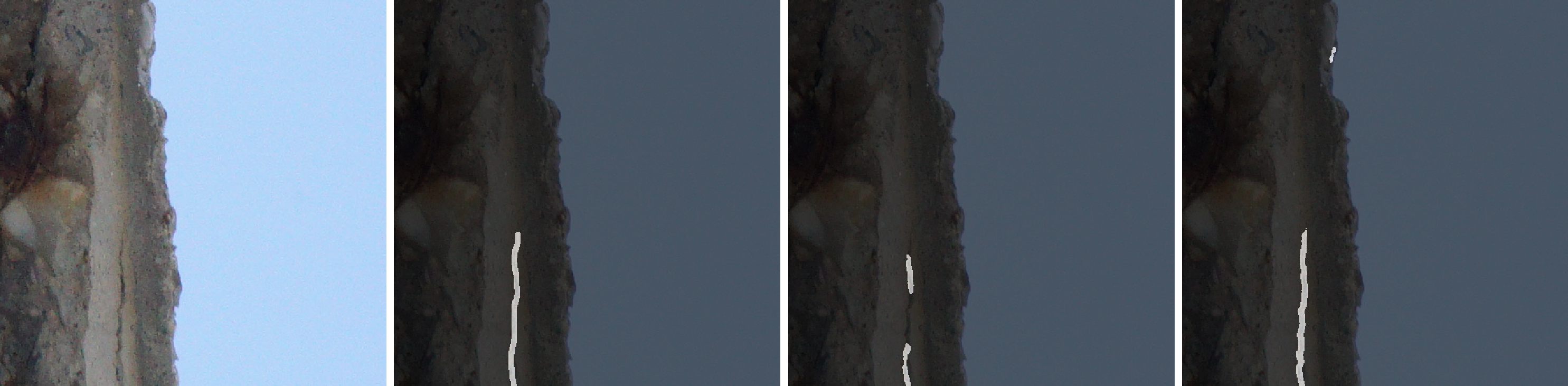}
	\caption{Qualitative examples on SegCODEBRIM from left to right: input image, ground-truth, U-Net, CAP-Net (ours). Images are compressed for view in pdf. Further examples can be found in the appendix.}
	\label{realExamples}
\end{figure*}

\begin{table*}[htb]
  \centering
  \begin{tabular}{@{}llcccccc@{}}
  & Model & $F1 (\uparrow)$ & $F1_{\theta=10} (\uparrow)$  & $clDice (\uparrow)$  & $HDF_{Euc} (\downarrow)$ & $HDF_{RBF} (\downarrow)$ \\ 
    \midrule
    \parbox[t]{2mm}{\multirow{6}{*}{\rotatebox[origin=c]{90}{SegCODEBRIM}}} & U-Net       &   { $29.5 \pm 3.1$} &    { $35.7 \pm 1.8$}  & { $38.5 \pm 0.9$}  &  { $40.2 \pm 13.2$} & { $53.5 \pm 7.0$} \\
& U-Net (PMI)       &   { $28.4 \pm 1.6$} &    { $30.7 \pm  2.5$}  & { $44.8 \pm 1.9$} &    { $31.2 \pm 2.1$}  &  { $53.3 \pm 2.7$} \\

  &      U-Net (AdaIn)       &   { $31.4 \pm 4.2$} &    { $33.8 \pm  4.3$}  & { $48.4 \pm 5.1$} &    { $26.4 \pm 5.2$}  &  { $48.5 \pm 6.6$} \\
   & CAP-Net w/o CL      &   { $32.6 \pm 2.8$} &    { $35.4 \pm 1.3$}  & { $48.7 \pm 1.8$} &     { $23.8 \pm 11.5$} & { $46.0 \pm 1.2$} \\

& CAP-Net w/o AdaIn       &   { $31.9 \pm 4.1$} &    { $40.1 \pm 1.9$}  & { $40.0 \pm 1.7$} &   { $41.0 \pm 14.7$} & { $52.2 \pm 0.8$} \\

 & CAP-Net &   {\boldmath $37.3 \pm 1.5$} &    {\boldmath $40.4 \pm 1.8$}  & {\boldmath $53.6 \pm 1.5$} &  {\boldmath $21.7 \pm 1.0$} & {\boldmath $40.4 \pm 1.8$} \\
    \bottomrule
  \end{tabular}
  \begin{tabular}{@{}llcccccc@{}}
     \parbox[t]{2mm}{\multirow{6}{*}{\rotatebox[origin=c]{90}{Multi-source Set}}} &  U-Net       &   { $42.6 \pm 0.2$} &    { $44.5 \pm 0.2$}  &   { $71.4 \pm 0.5$}  &  { $13.7 \pm 1.3$} & { $33.5 \pm 1.0$} \\
& U-Net (PMI)       &   { $40.6 \pm 0.9$} &    { $42.4 \pm 1.1$}   &    { $68.4 \pm 0.5$}  &  { $18.3 \pm 0.5$} & { $37.8 \pm 1.1$} \\

      &  U-Net (AdaIn)       &   { $40.6 \pm 1.6$} &    { $42.0 \pm 1.5	$}  &    { $70.7 \pm1.2$}  &  { $14.1	 \pm 1.4$} & { $36.1 \pm 2.7$} \\

  &  CAP-Net w/o CL       &   { $42.0 \pm 8.1$} &    { $43.9 \pm 8.6$} &    { $69.5 \pm 7.7$}  &  { $14.4 \pm 3.9$} & { $34.6 \pm 9.1$} \\
   & CAP-Net  w/o  AdaIn    &   {\boldmath $45.2 \pm 2.4$} &    {\boldmath $47.1 \pm 1.6$}   &    {\boldmath $72.6 \pm 0.7$}  &  { \boldmath $13.1 \pm 0.6$} & { \boldmath $30.5 \pm 2.3$} \\
   & CAP-Net &   { \boldmath $44.9 \pm 1.5$} &    { \boldmath $46.9 \pm 1.8$}  & {\boldmath $72.3 \pm 1.5$} &  {\boldmath $13.2 \pm 1.0$} & { \boldmath $31.5 \pm 1.8$} \\
  \end{tabular}
  \smallskip
  \caption{Ablation study on SegCODEBRIM (top half) and multi-source set (bottom half). The best performing models on each dataset are highlighted in bold, where multiple values are highlighted if they lie within statistical deviations. Here, U-Net (PMI) and U-Net (AdaIn) refer to only the bottom and top parts of figure \ref{mainPipeline} respectively, whereas w/o CL and AdaIn denote the omission of the contrastive term and style transfer module. The ``full'' CAP-Net demonstrates that each designed component is crucial for SegCODEBRIM, whereas some components like AdaIn may be optional to perform well on the multi-source data.}
  \label{tableRes2}
\end{table*}

\noindent \textbf{(Q4) All CAP-Net design choices contribute to performance improvements:} 
The ablations in Table \ref{tableRes2}   shows the performance of different sub-modules of our system on SegCODEBRIM and multi-source set. 

First, the style transfer provided by AdaIN improves the generalization to real world data compared to a simple U-Net on SegCODEBRIM (U-Net(AdaIN) vs CAP-Net w/o AdaIn), but leads to statistically insignificant performance change on multi-source data. Second, the  addition of a second encoder branch (bottom half of figure \ref{mainPipeline}) that receives affinity scores based on PMI as input further increases the performance on most metrics, even when the branches are not contrasted (CAP-Net w/o CL). For instance, we obtain a $1.5 \%$ improvement in F1 and $3$ on the RBF Hausdorff distance on SegCODEBRIM. Third, the subsequent addition of the consistency loss leads to consolidated segmentation maps between both encoders and improves performance on various metrics (CAP-Net w/o CL vs. ``full'' CAP-Net). We obtain a $5\%$ improvement in F1 and decrease of $6$ in Hausdorff distance respectively on SegCODEBRIM. 

The results of Table \ref{tableRes2} empirically corroborate the efficacy of our design choices. The incorporation of PMI-based modeling approaches and purely data-driven U-Net style learning, augmented by consistency loss improves the appearance invariance of our model and thus leads to better generalization to out of distribution data.






\section{Conclusion}

In this paper, we introduced Cracktal, a flexible simulator for generating synthetic cracked concrete surface data with ground truth labels.  Additionally, we proposed a hybrid design that combines data-driven models with single-image statistical estimation models, to fully leverage synthetic data. Our empirical validation demonstrates that this approach reduces the Sim2Real gap. Our work emphasizes the importance of fusing expert-based inductive biases with learning from simulated data and provide new domain to explore domain generalization and adaptation methods. 

\section{Acknowledgments}

We acknowledge funding from the European Union H2020  Research and Innovation Programme under grant agreement number 769066. This work was also supported by the Artificial Intelligence Systems Engineering Laboratory (AISEL) project under funding number 01IS19062, funded by the German Federal Ministry of
Education and Research (BMBF) program "Einrichtung von KI-Laboren zur Qualifizierung im Rahmen von Forschungsvorhaben im Gebiet der Künstlichen Intelligenz".


{\small
\bibliographystyle{ieee_fullname}
\bibliography{egbib}
}

\end{document}


\title{Appendix: Designing a Hybrid Neural System to Learn Real-world \\ Crack Segmentation from Fractal-based Simulation}

\maketitle

\appendix 

In this supplementary material, we provide additional details, experimental setup and descriptions for the various parts  of our proposed systems (Cracktal and CAP-Net), as well as discuss additional results and highlight the limitations of our approach. The structure is as follows:

\begin{itemize}
    \item[\textbf{A.}] Experimental setup and implementation details of the Cracktal Simulator.
    \item[\textbf{B.}] Additional training and design details of CAP-Net.
    \item[\textbf{C.}] Additional results on in-distribution Cracktal data.
    \item[\textbf{D.}] Further qualitative results on real world images.
    \item[\textbf{E.}] Discussion of the limitations of our current approach and possible future improvements.
\end{itemize}

\section{Cracktal Details}
In this section, we discuss some fundamentals of the physics based workflow and the texture maps used in Cracktal. We also list the hyperparameters that have been used to generate our specific Cracktal data set.
\subsection{Rendering Maps and Crack Simulation Model}
\begin{figure}[htbp]
	\centering
	\includegraphics[width=0.32\linewidth]{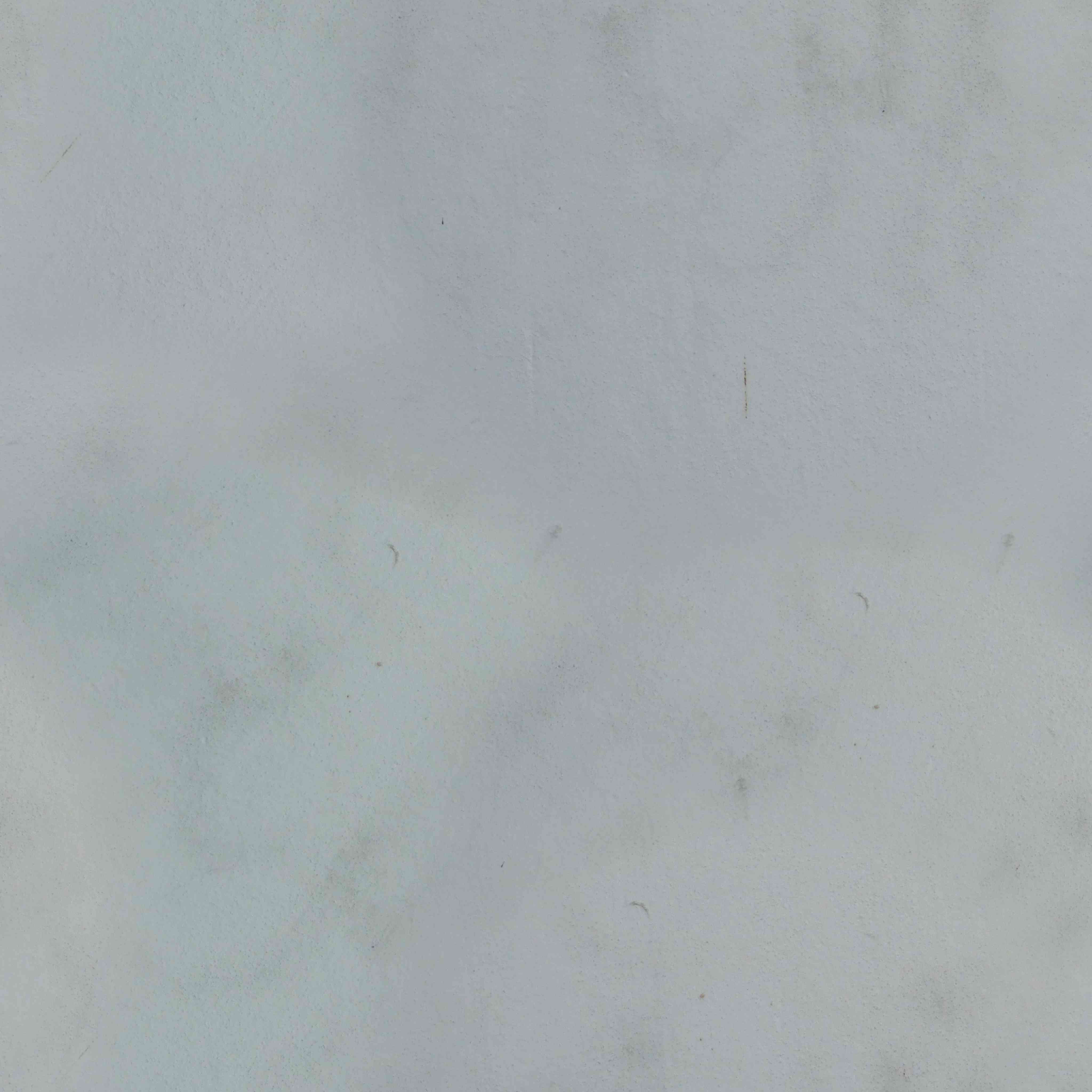}
\includegraphics[width=0.32\linewidth]{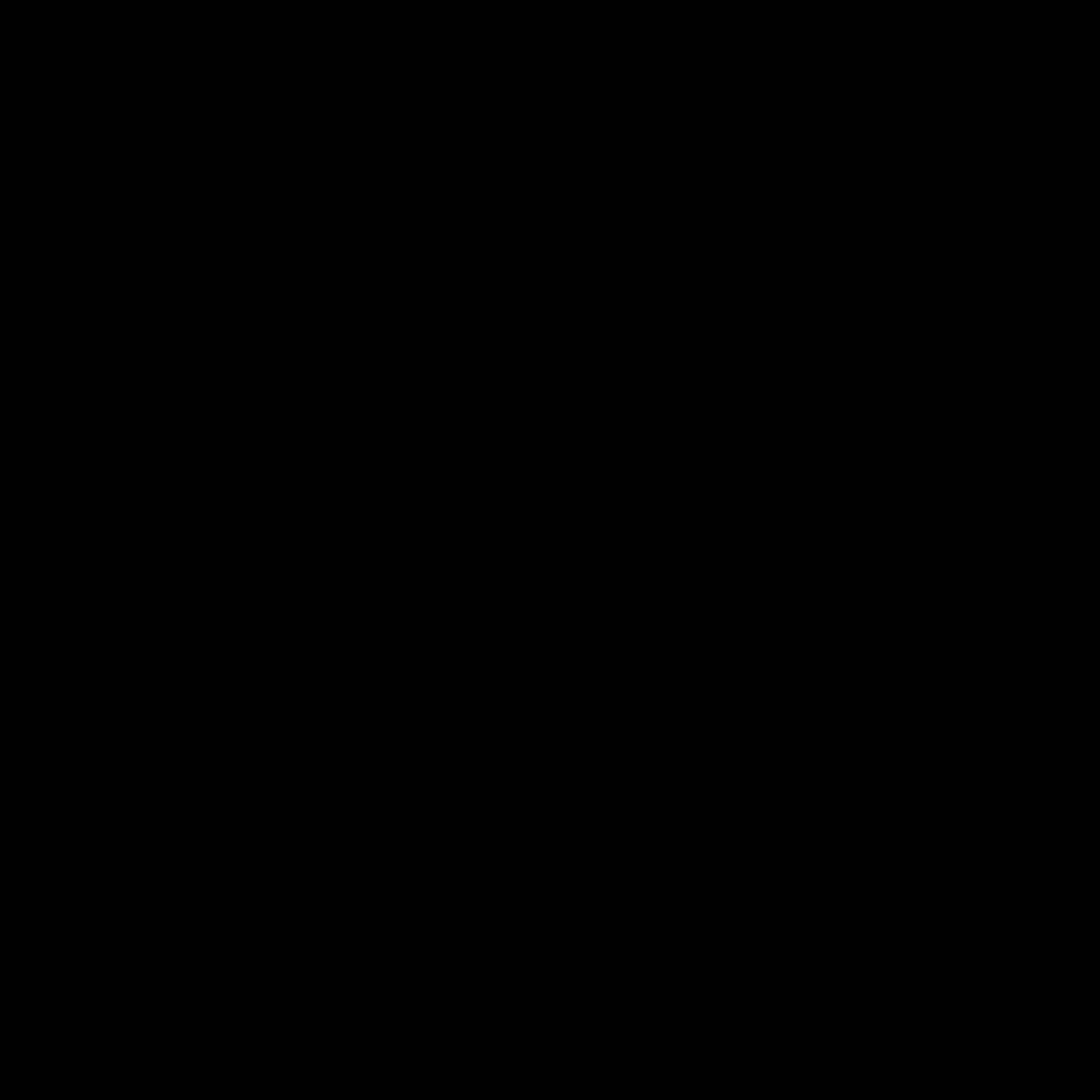}
\includegraphics[width=0.32\linewidth]{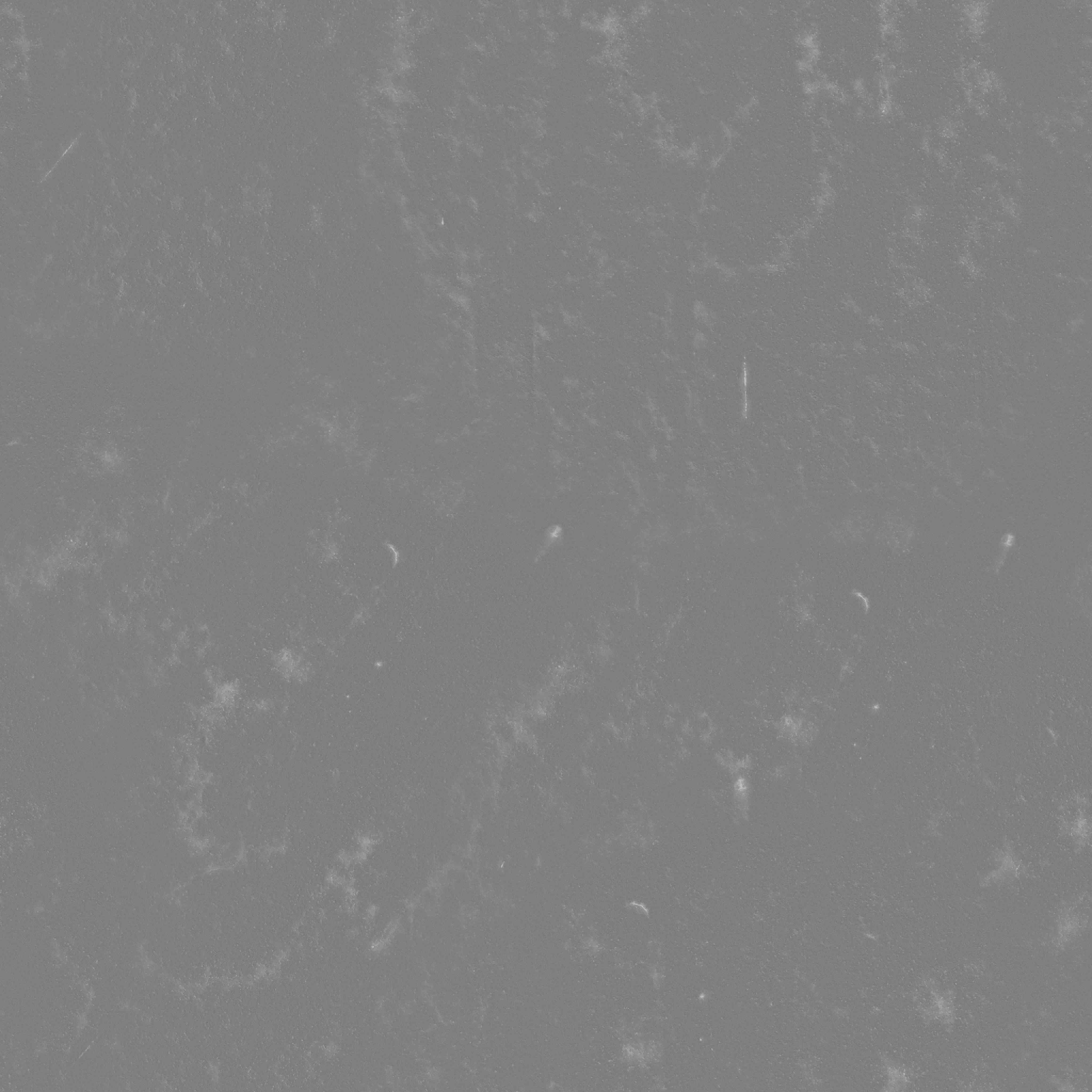}
\includegraphics[width=0.32\linewidth]{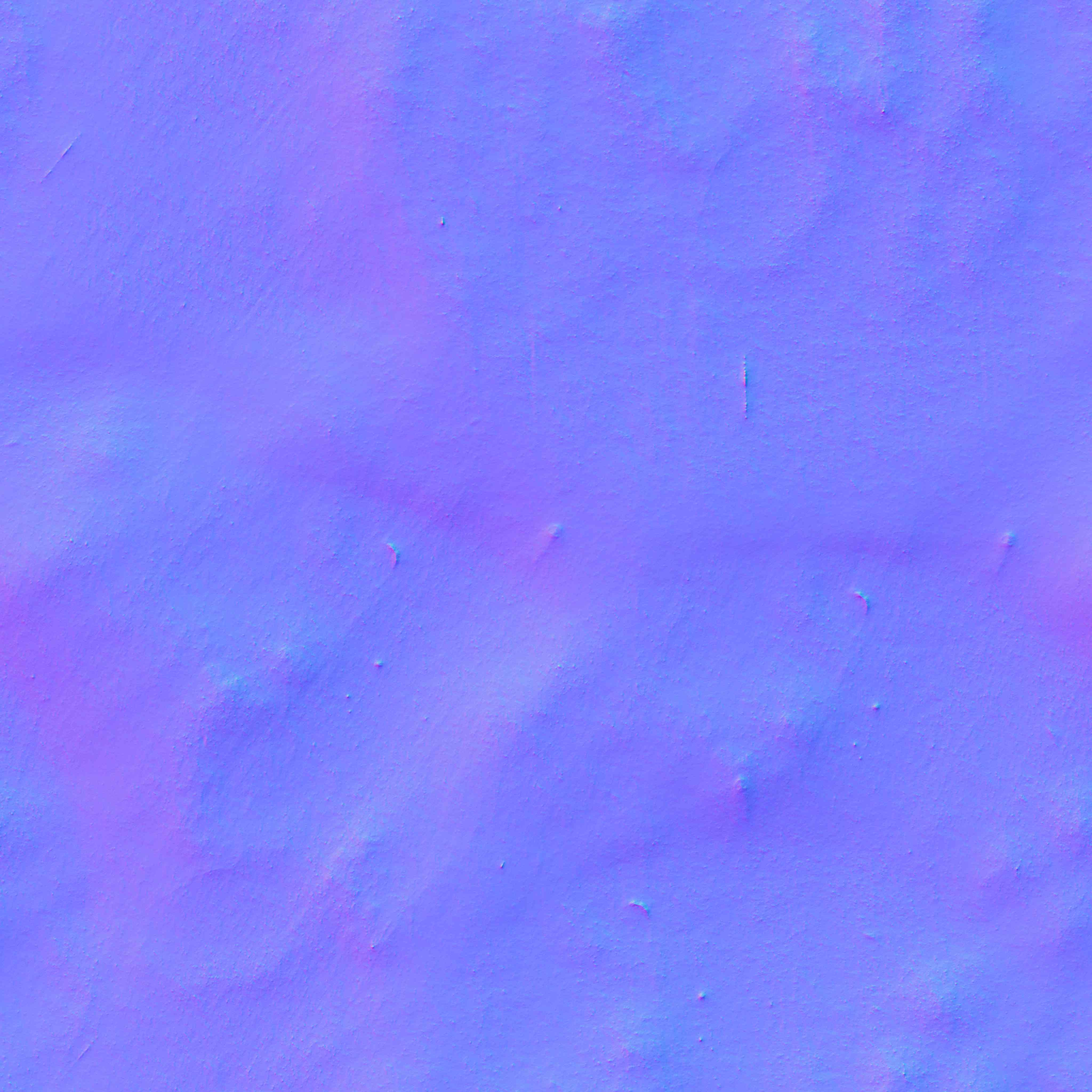}
\includegraphics[width=0.32\linewidth]{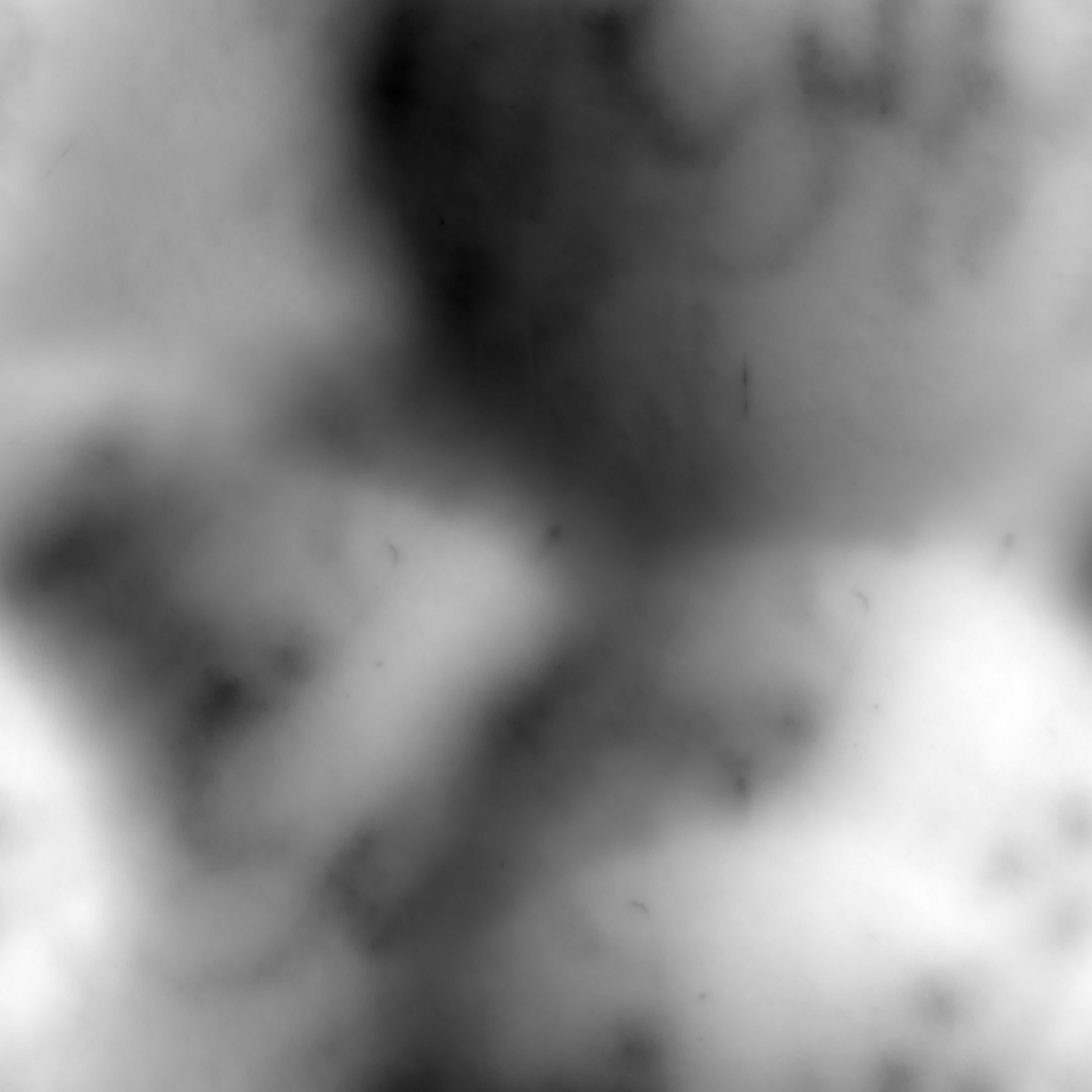}
\includegraphics[width=0.32\linewidth]{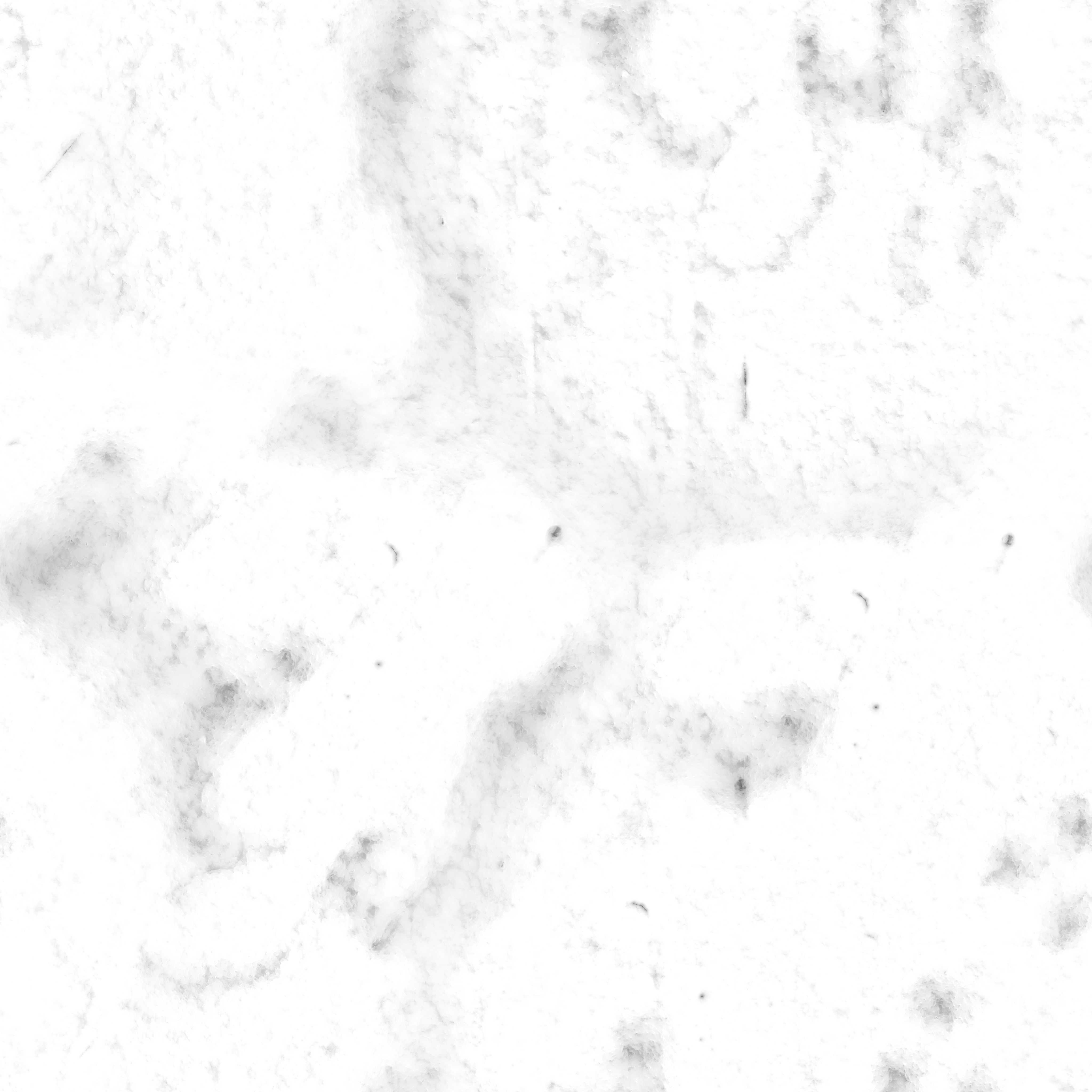}
	\caption{PBR Maps used to generate a concrete surface. Textures were compressed to view in PDF. From left to right, top to bottom: albedo, metallic, roughness, surface normal, height, and ambient occlusion map.}
	\label{PBR_textures}
\end{figure}

\noindent \textbf{PBR Maps:} Physics-based rendering (PBR) is an approach to rendering scenes that aims to accurately simulate the behavior of light in the real world. In PBR, materials are defined by their physical properties, such as color (in the case of dielectrics), roughness, and surface normals, which are used to calculate how light interacts with the material. In practice, PBR relies on a set of texture maps that define the physical properties of each object in the scene (see Figure \ref{PBR_textures}). These texture maps include:

\begin{itemize}
    \item Color/albedo map: This texture map defines the base color of the material for our case of the fully dielectric concrete (or alternatively reflectance value for metals), which represents the color that is observed under diffuse lighting conditions.
    \item Metallic Map: This map defines the metallicity of a material. Values range from non-metallic (zero values) to fully metallic (one values), and similarly to a mask, define how to treat the above color map's values in specific regions with respect to reflected color or specularity.   
    \item Roughness Map: This map defines the smoothness of the material, with values ranging from very rough to very smooth. Whereas the amount of reflected light is always conserved, roughness determines diffuse scattering. That is, the reflected direction becomes increasingly random the rougher the surface. 
    \item Normal Map: This map encodes the surface normals of a material in RGB, defining which direction a surface faces and thus affecting light and shadow calculations. Normal maps are typically created by capturing fine surface details through photogrammetry.
    \item  Height map: This map is used to modify the surface geometry. It defines the height variations of the surface of an object in a grayscale map, where brighter values represent higher elevations and darker values represent lower ones. It is used in conjunction with other material properties, such as roughness and surface normal maps, to calculate how light interacts with the surface of an object and providing a more realistic ``bumpmapped'' look. 
    \item Ambient Occlusion Map: This map defines the occluded areas on a surface, which are areas that receive less light due to being blocked or shadowed by other objects. AO maps help to add depth and realism to a material by accentuating the small details and crevices in a surface.
\end{itemize}

By using these texture maps in a PBR renderer, it is possible to create highly realistic materials that look similar to their real-world counterparts under different lighting conditions. For more detailed information, we defer to popular PBR guides, such as by Adobe's Substance\footnote{see \url{https://substance3d.adobe.com/tutorials/courses/the-pbr-guide-part-2} PBR Guide 2018}. \\

\begin{figure}[htbp]
	\centering
	\includegraphics[width=0.32\linewidth]{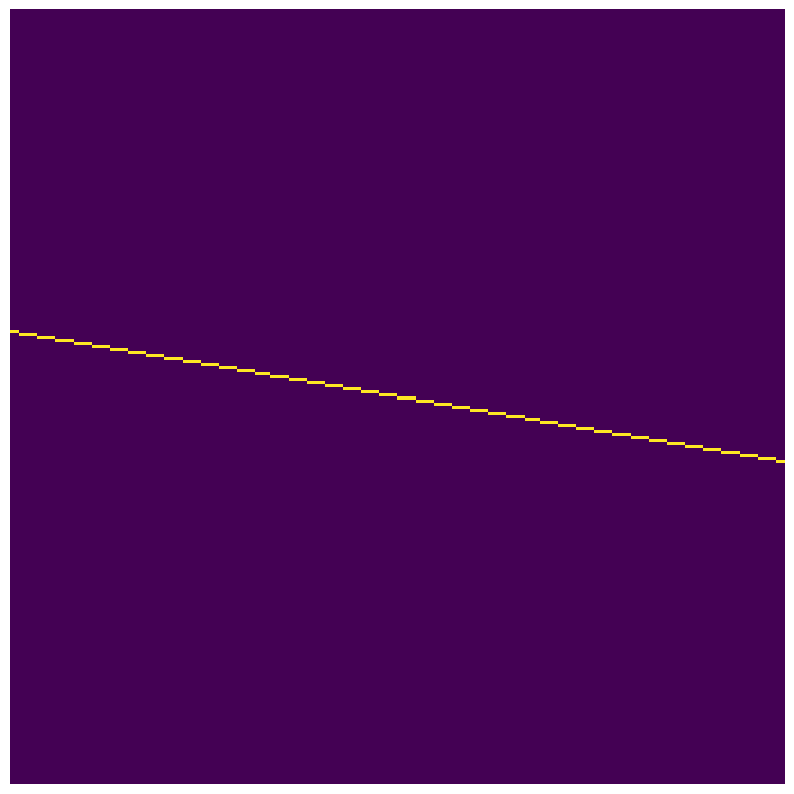}
\includegraphics[width=0.32\linewidth]{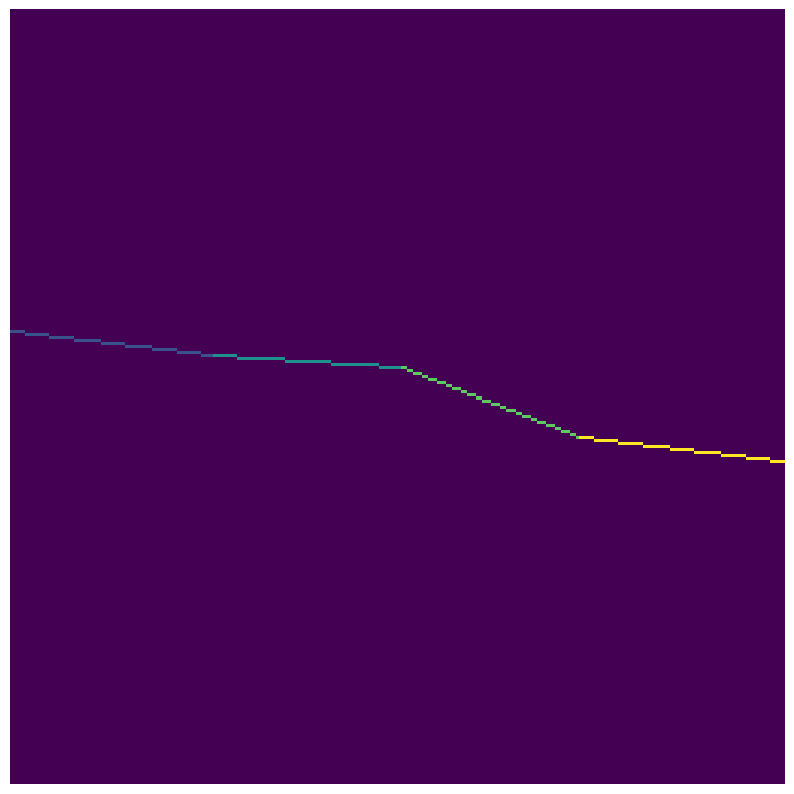}
\includegraphics[width=0.32\linewidth]{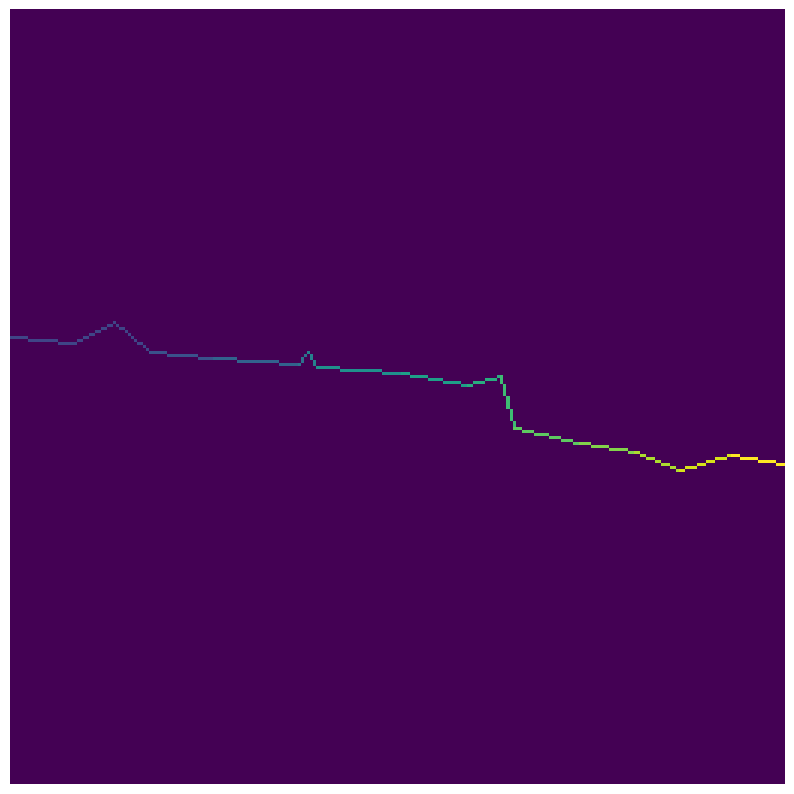}
\includegraphics[width=0.32\linewidth]{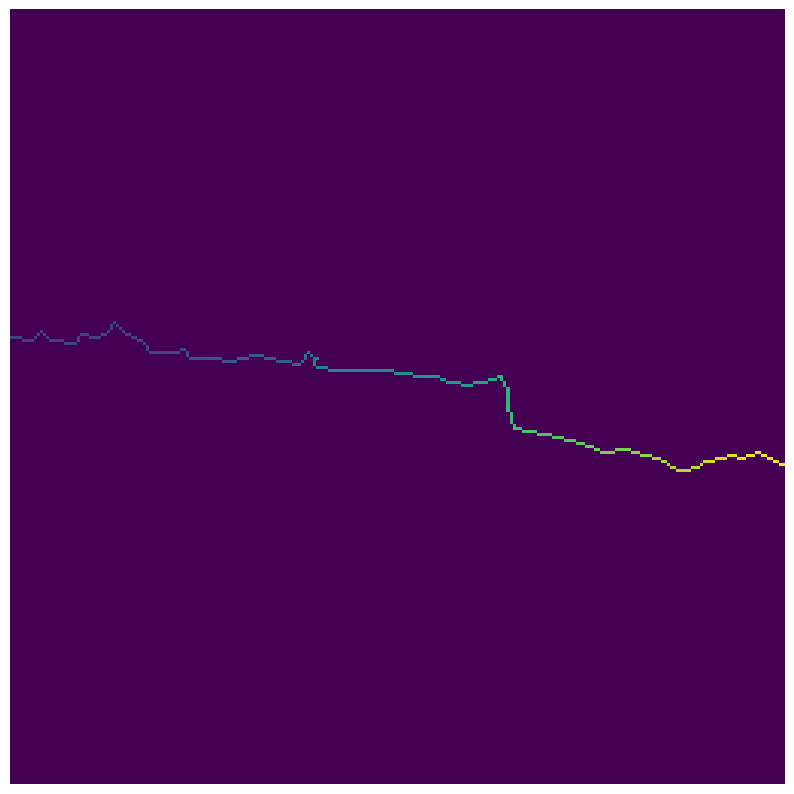}
\includegraphics[width=0.32\linewidth]{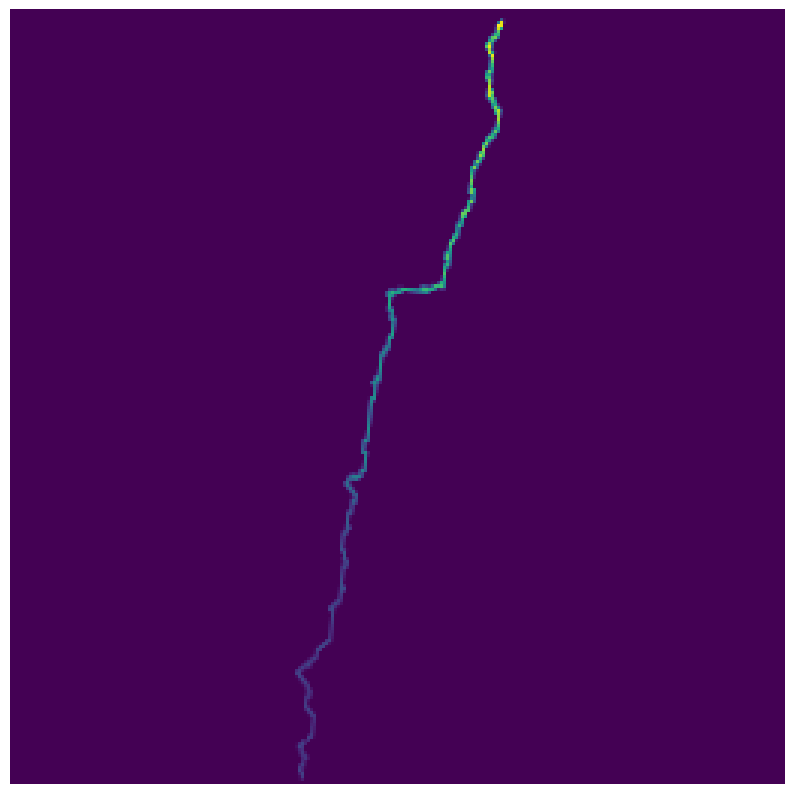}
\includegraphics[width=0.32\linewidth]{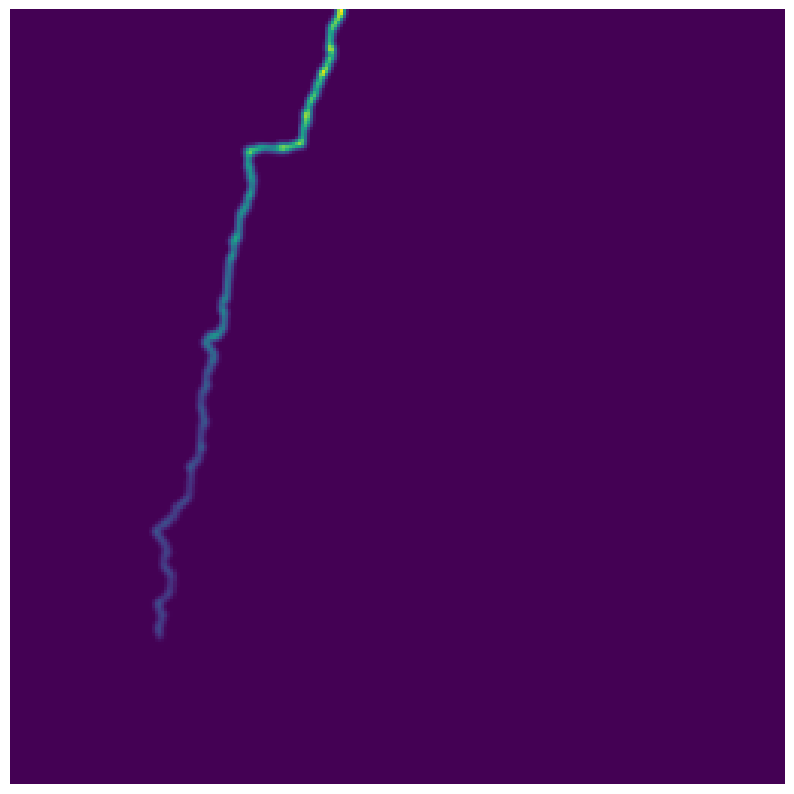}
	\caption{Crack Generation steps. A straight line is modified iteratively using the stochastic midpoint displacement algorithm. In the final steps the crack is translated, rotated and width is then added by applying a Gaussian blur filter. }
	\label{FractalExample}
\end{figure}

\noindent \textbf{Crack Model:}
As explained in the main body, the crack model is implemented using an irregular fractal model. In our experiments, the subdivision depth is set to 7. Furthermore, the width of the crack is obtained by  generating a random kernel size (3 or 5) for a Gaussian blur and applying it to the crack map. Figure \ref{FractalExample} illustrates the crack generation process. The obtained crack is then added to the scene.  

\subsection{Implementation Details}
We generate our data using  Blender Cycles PBR engine.
Our simulator code is written using Blender's Python API, which provides a comprehensive set of functions and classes for interacting with the Cycles Engine.  The code was scripted in blender version 2.79 on a Linux OS.
We will open source the Cracktal code, the underlying textures, as well as the generated data for training and validation. With respect to surface alterations, the employed moss textures were downloaded from textures.com \footnote{\url{https://www.textures.com/search?q=moss}} and the graffiti textures from turbosquid.com \footnote{\url{https://www.turbosquid.com/FullPreview/490921}}. \\

\noindent \textbf{Rendering Hyperparameters:} We set the sampling factor for the data generation to 20. When rendering an image, the rendering engine sends out rays from the camera into the scene and estimates the color of the first object that each specific ray intersects with. To reduce aliasing and noise, the renderer takes multiple samples per pixel and averages the estimated colors to obtain a more accurate representation of each pixel in the scene. Generally, higher sampling rates lead to more accurate and realistic results but also require more computational resources and time. We chose a rendering tile size of $256 \times 256$ (for parallelization) and a resolution of $2048 \times 2048$ for the final image. \\

\section{CAP-Net}
In this section, we present further details and visualization of the main modules in our proposed system, CAP-Net. We also document our training and validation procedure.  
\subsection{Pointwise mutual information}
\begin{figure}[htbp]
	\centering
	\includegraphics[width=0.32\linewidth]{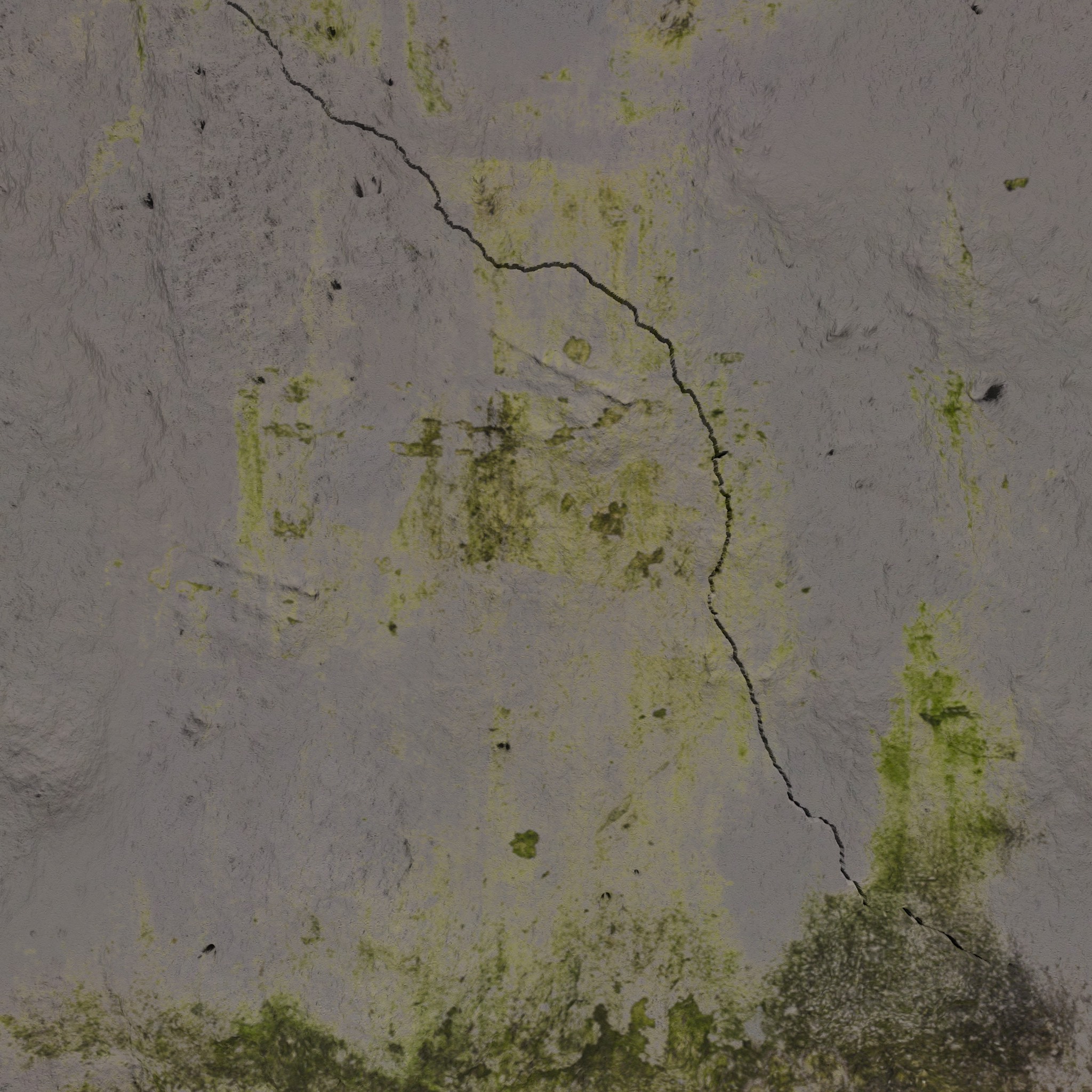}
\includegraphics[width=0.32\linewidth]{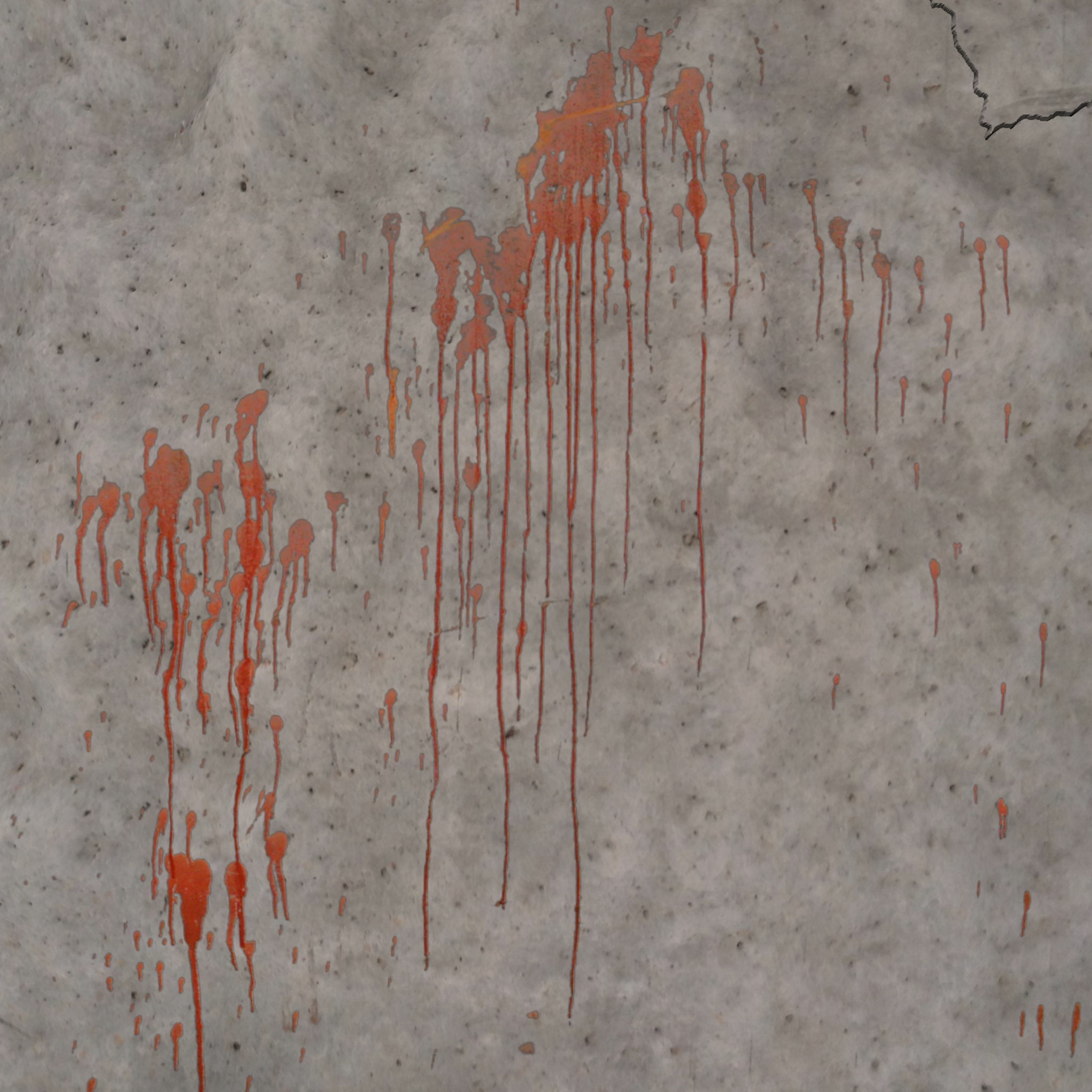}
\includegraphics[width=0.32\linewidth]{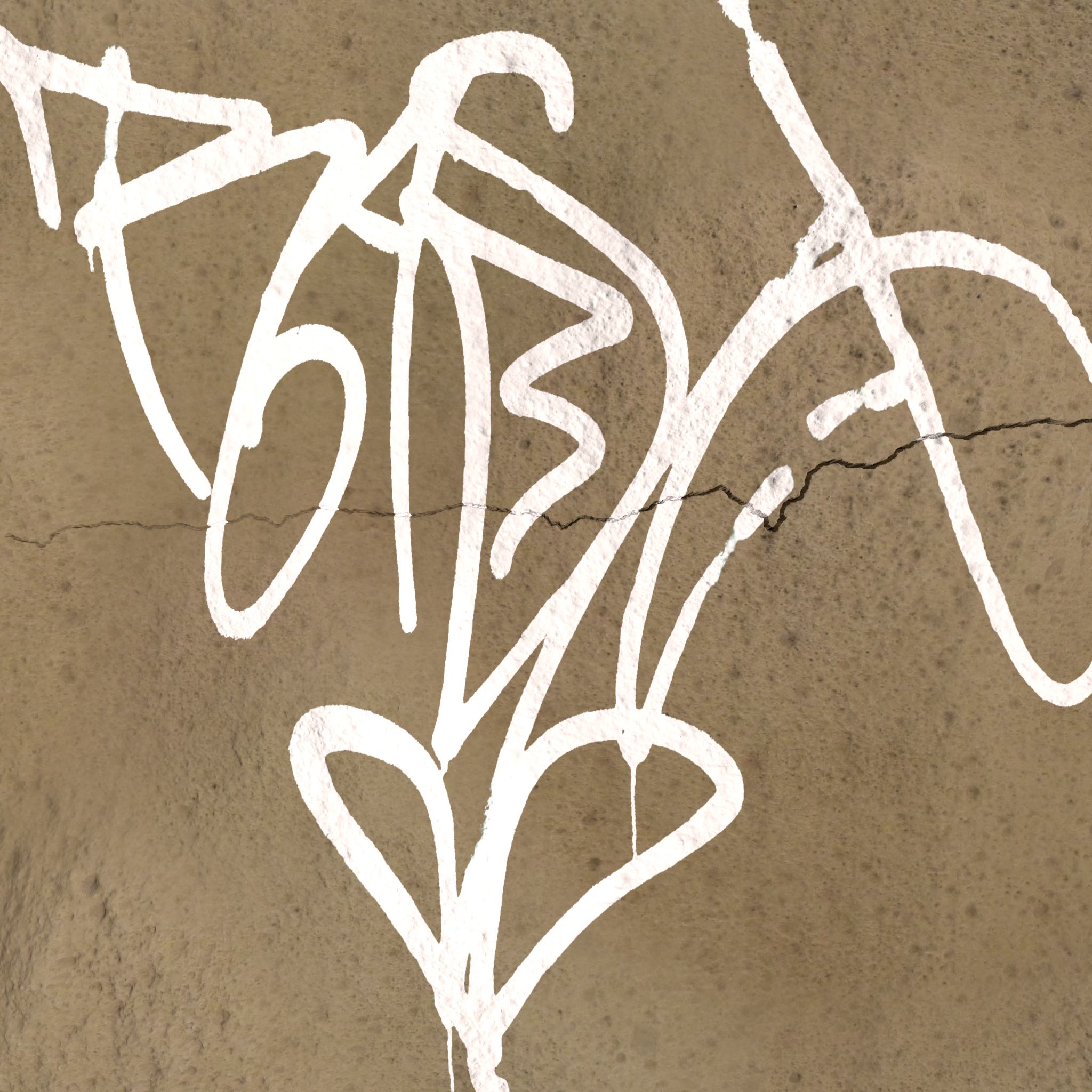} \\
\includegraphics[width=0.32\linewidth]{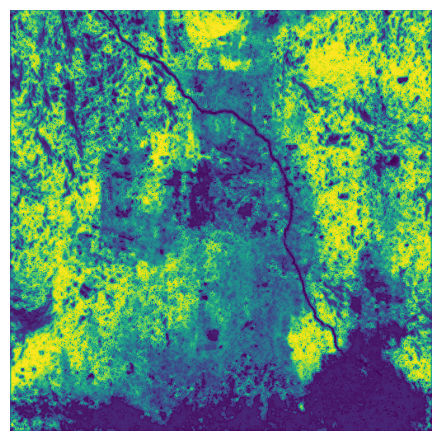}
\includegraphics[width=0.32\linewidth]{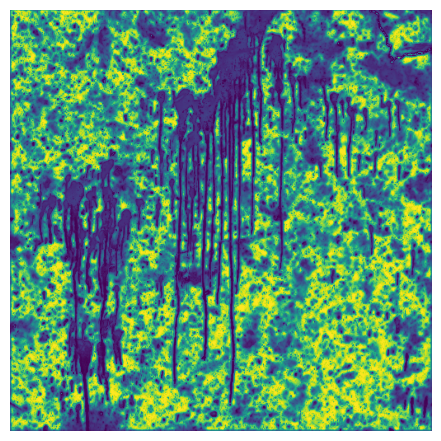}
\includegraphics[width=0.32\linewidth]{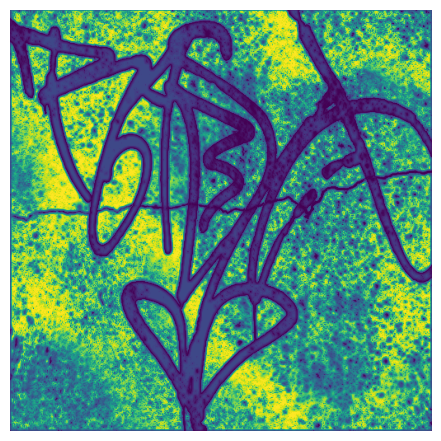}
	\caption{Examples of the affinity maps obtained by the pointwise mutual information module.  Images are compressed for view in PDF. Upper row shows the original images from Cracktal dataset and the lower row is their corresponding affinity maps.}
	\label{Pmiexamples}
\end{figure}

The basic assumption underlying the choice of a pointwise mutual information module as an inductive bias is that the statistical association between pixels belonging to the background texture is high, whereas for pixels belonging to anomalies their statistical association with neighboring pixels is low. We hypothesize that by leveraging these statistical associations between neighboring pixels for learning, we can achieve better generalization from simulated to real images, which is empirically corroborated in the main body. Figure \ref{Pmiexamples} presents some visualization of the obtained pixel wise affinity scores. Here we see that cracks and other anomalies like surface holes, moss or graffiti edges are assigned lower scores. 

\subsection{Adaptive Instance Normalization}
\begin{figure}[htbp]
	\centering
	\includegraphics[width=0.32\linewidth]{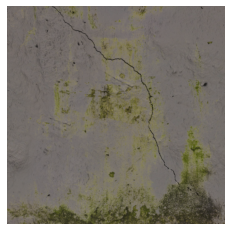}
\includegraphics[width=0.32\linewidth]{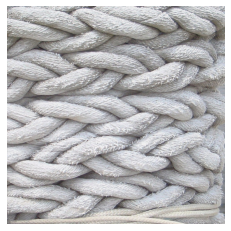}
\includegraphics[width=0.32\linewidth]{/igures_appendix/adain_examples/res1.png} \\
\includegraphics[width=0.32\linewidth]{figures_appendix/adain_examples/orig.png}
\includegraphics[width=0.32\linewidth]{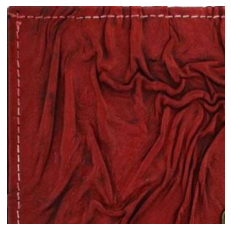}
\includegraphics[width=0.32\linewidth]{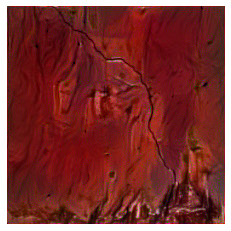} \\

\includegraphics[width=0.32\linewidth]{figures_appendix/adain_examples/orig.png}
\includegraphics[width=0.32\linewidth]{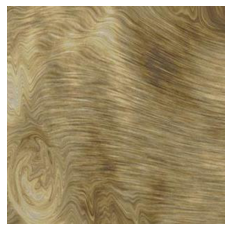}
\includegraphics[width=0.32\linewidth]{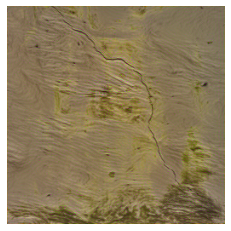}
	\caption{Adaptive Instance Normalization examples. Images are compressed for view in PDF. From left to right: original, style, and stylized images.}
	\label{adainexamples}
\end{figure}

Figure \ref{adainexamples} visualizes some examples of images transformed by the AdaIN module. We use a similar design to that of the original AdaIn work \cite{huang2017arbitrary}. The style transfer network takes a content image c and an arbitrary style image s as inputs, and synthesizes an output image that recombines the content and the style. A simple encoder-decoder architecture is used, in which the encoder f is fixed to the first four layers of VGG-19 pretrained on ImageNet. After encoding the content c and style s images to the latent feature space, both feature maps are passed to  an AdaIN layer that aligns the mean and variance of the content feature map. A randomly initialized decoder is trained to map the obtained features to the image space, generating the stylized image. The loss function of the decoder is computed as follows: 
\begin{align}
    \pazocal{L} = \lambda \cdot \pazocal{L}_s + \pazocal{L}_c
\end{align}

which is a weighted combination of the content loss $\pazocal{L}_c$ and the style loss $\pazocal{L}_s$ with the style loss weight $\lambda$. The content loss term $\pazocal{L}_c$ is computed as the Euclidean distance between the target features and the features of the output image. The style loss $\pazocal{L}_s$ is the Gram matrix loss between the decoded image and the style image s.

We set the style weight $\lambda=10000$ and perform $1000$ iteration steps to generate a stylized image. For each image in our training set, we generate $5$ stylized images and pick randomly between them during training. As an optimizer for the decoder, we use LBFGS \cite{liu1989limited}. As observable in figure \ref{adainexamples}, the background texture is changed to match the style of the style image s, but the crack remains visible. 

\subsection{Implementation Details}
In our experiments, we use a U-Net \cite{ronneberger2015u} architecture as the backbone of our semantic segmentation network. The encoders are implemented in PyTorch and trained for 30 epochs on two A100 Nvidia GPUs using Adam with an initial learning rate of $0.001$ and weight decay of $0.0005$, mini-batch size 32. We train the models using synthetic concrete images generated with our Cracktal simulator. We save the model with the highest validation accuracy on a validation set of synthetic images. 

During training, we downsample the Cracktal images to a resolution of $512 \times 512$ and then randomly crop a $256 \times 256$ patch. We don't perform any data augmentation during training. The training set is composed of 8800 images, as this amount seems to saturate performance (see section C). 

For evaluation on SegCODEBRIM, we process and segment the original images with resolution $1500 \times 844$ patch-wise. Each patch has a resolution of $512 \times 512$ and is then downsampled to $256 \times 256$ to input to the neural network.

For the pointwise mutual information estimation, we sample $10000$ pixel pairs to perform the kernel density estimation, which is a good computational trade-off given that the input images have a $256 \times 256$ resolution. 
The distance between two pixels in a pair is between 1 and 8 pixels.  As a kernel density estimator, we use gaussian KDE with automatic bandwidth estimation \cite{scott2015multivariate}. For each pixel, we estimate the affinity by calculating the PMI scores with its neighbouring pixels in a radius of 5 from the origin. As a hyperparamater for the PMI, we set $\tau=2.25$.

\section{Additional Results}
\begin{table*}[t]
  \centering
  \begin{tabular}{@{}llcccccc@{}}
     & Model &$F1 (\uparrow)$ & $F1_{\theta=10} (\uparrow)$  & $clDice (\uparrow)$  & $HDF_{Euc} (\downarrow)$ & $HDF_{RBF} (\downarrow)$ \\ 
    \midrule
   \parbox[t]{2mm}{\multirow{6}{*}{\rotatebox[origin=c]{90}{Cracktal}}} & U-Net  \cite{ronneberger2015u}     &   { $75.3 \pm 0.7$} &    { $82.4 \pm 0.2$}  & { $88.2\pm 0.4$}  &  { $11.0 \pm 0.9$} & { $15.3 \pm 0.2$} \\
   & Attn-U-Net \cite{oktay2018attention}      &   { $73.3 \pm 0.6$} &    { $81.4 \pm 0.1$}  & { $86.3 \pm  0.7$}  &  { $13.6 \pm 1.1$} & { $17.6 \pm 0.7$} \\
    
   & Multi-U-Net (D-SN)     &   { $73.2 \pm 1.1$} &    { $81.4 \pm 0.8$}  & { $84.2 \pm 1.5$}  &  { $13.6 \pm 3.1$} & { $14.5 \pm 2.7$} \\

  &  Multi-U-Net (PMI)      &   { $72.8 \pm 4.1$} &    { $80.4 \pm 3.1$}  & { $83.2 \pm 7.3$}  &  { $14.3 \pm 5.2$} & { $18.5 \pm 6.1$} \\

  &  CAP-Net &   { $72.2 \pm 0.9$} &    { $80.1 \pm 0.4$}  & { $86.9 \pm 1.3$} &  { $11.7 \pm 1.5$} & { $17.6 \pm 0.8$} \\
    
    \bottomrule
  \end{tabular} \smallskip
  \caption{Performance comparison of different models on a testset of Cracktal data.}
  \label{cracktalLabel}

\end{table*}

\begin{figure*}[htbp]
	\centering
 \includegraphics[width=0.33\linewidth]{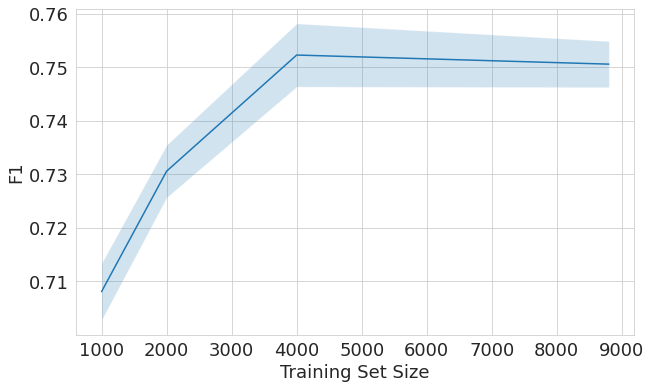}
 \includegraphics[width=0.33\linewidth]{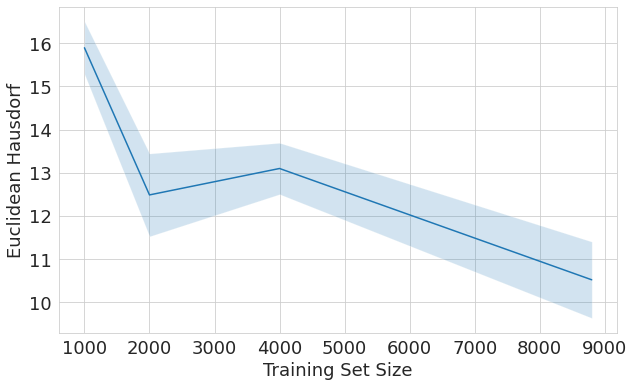}
 \includegraphics[width=0.33\linewidth]{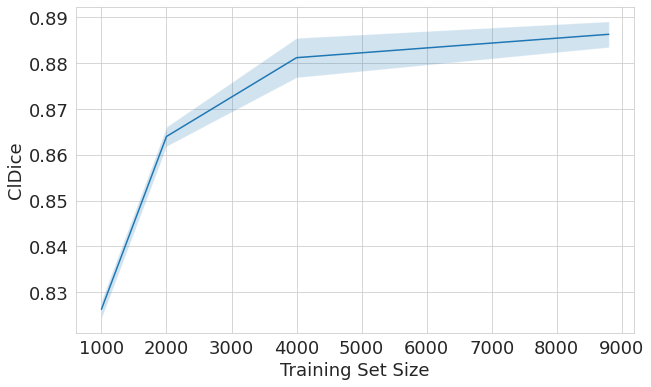}
	\caption{Evolution of segmentation performance in terms of training set size for baseline U-Net on Cracktal testset. We visualize the performance using different metrics. From left to right: F1,  Hausdorff distance with euclidean measure, clDice.}
	\label{PlotsSetsize}
\end{figure*}

\begin{figure*}[htbp]
	\centering

 \includegraphics[width=\linewidth]{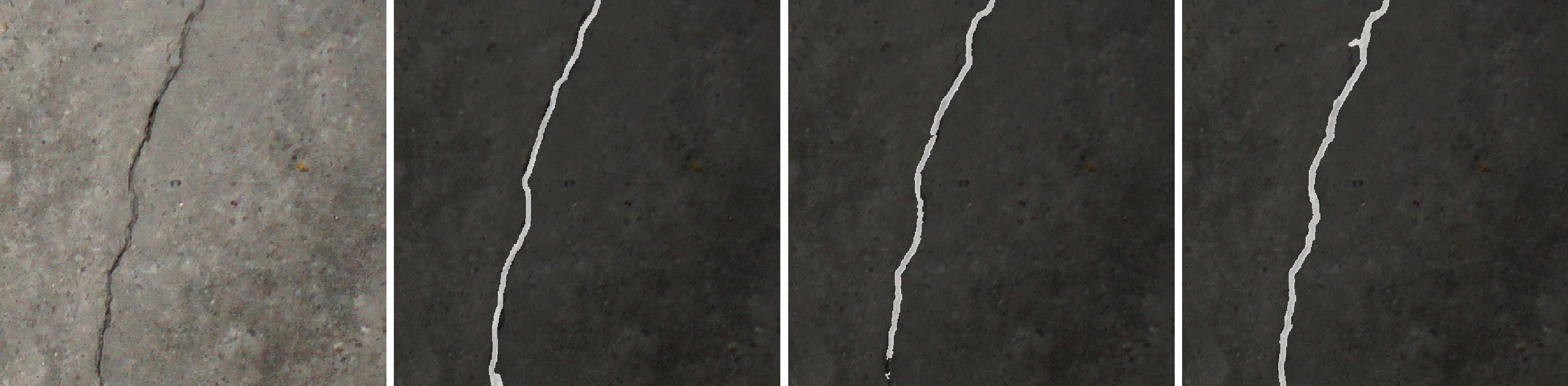}
 \\

  \includegraphics[width=\linewidth]{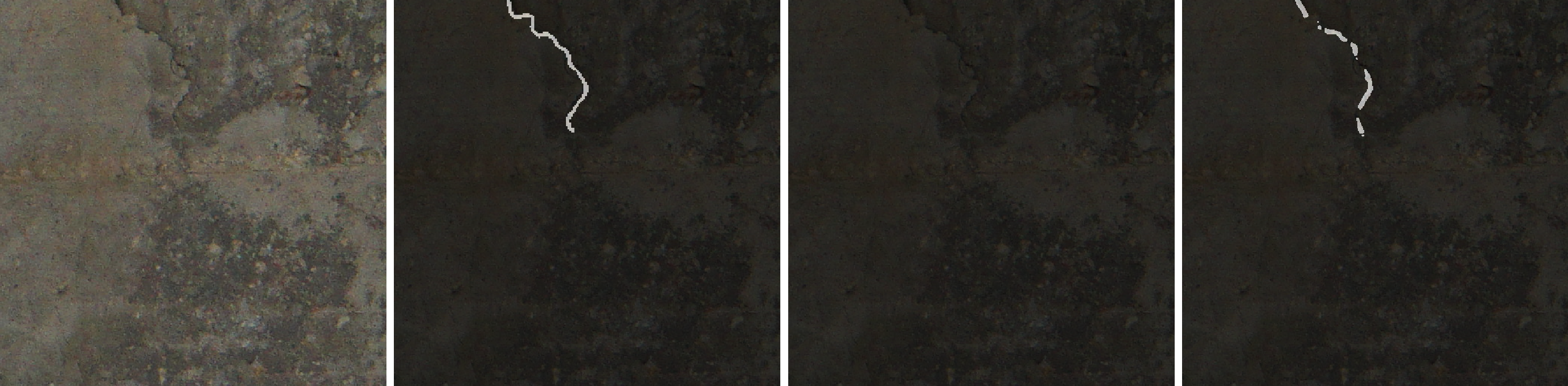} \\

 \includegraphics[width=\linewidth]{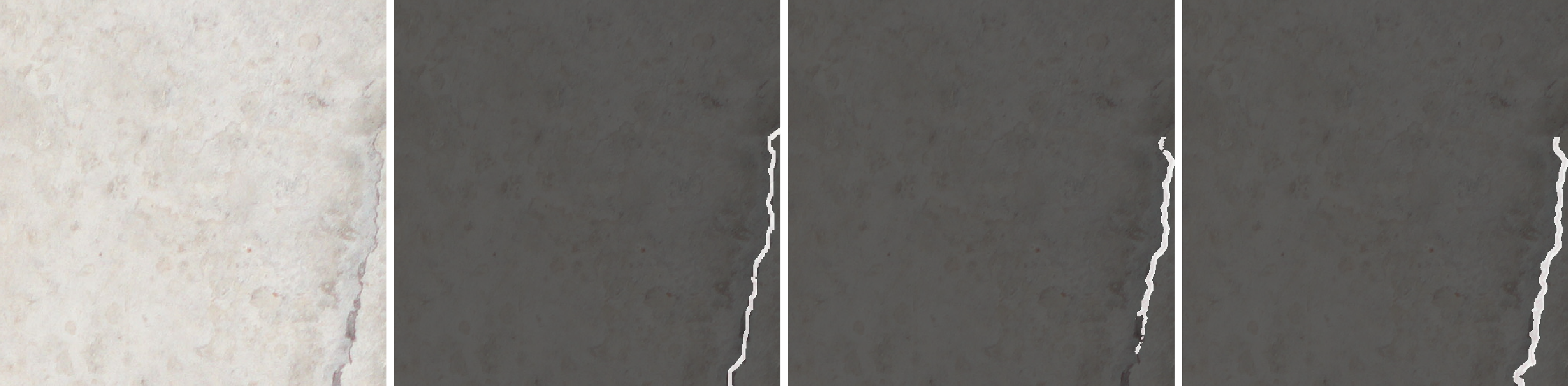}
 \\

 \includegraphics[width=\linewidth]{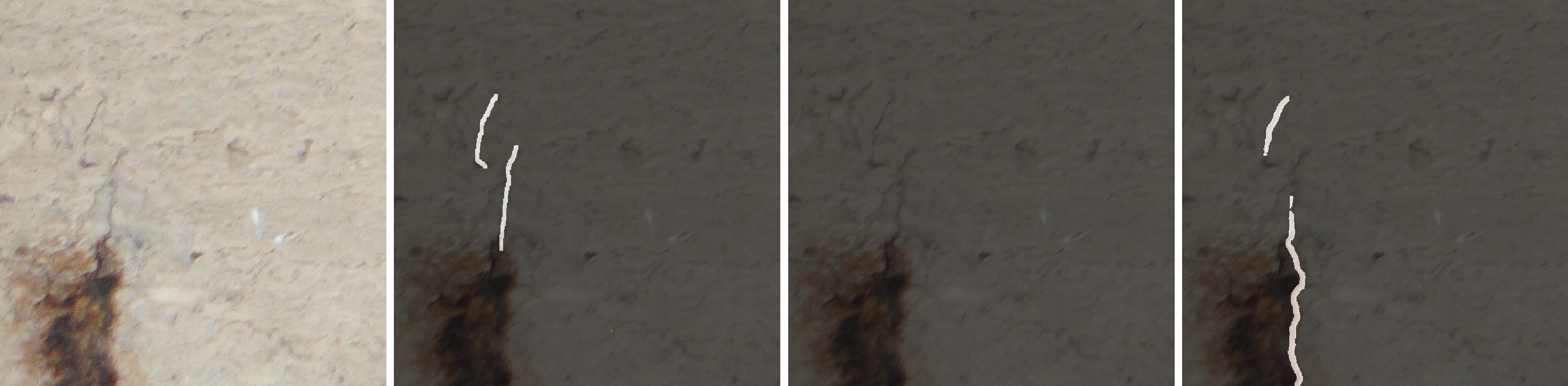}
 \\

 \includegraphics[width=\linewidth]{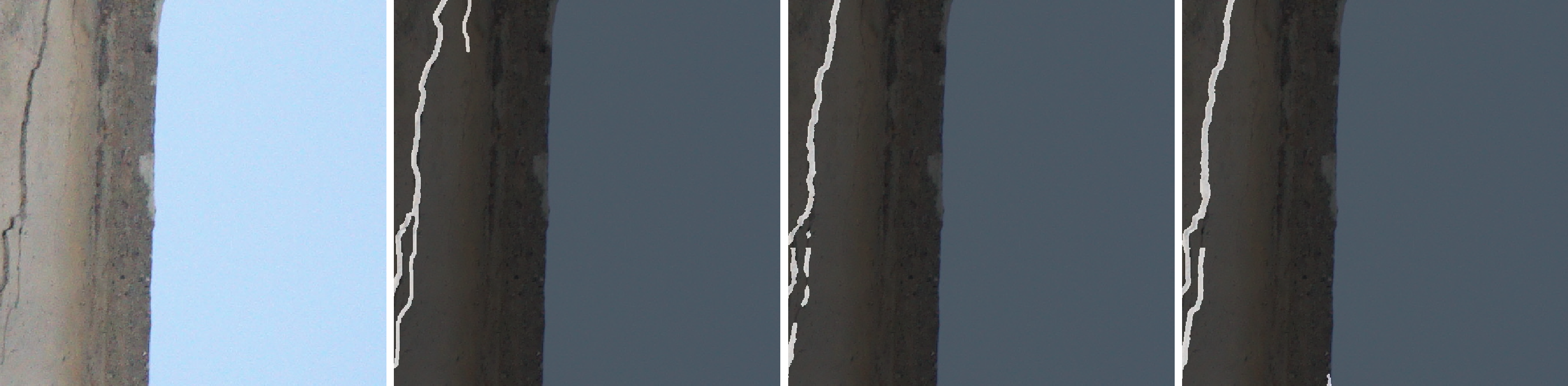}
 \\
	\caption{Qualitative examples on SegCODEBRIM from left to right: input image, ground-truth, U-Net, CAP-Net (ours). In general, CAP-Net segments cracks that U-Net often misses entirely. Images are compressed for view in pdf. }
	\label{realExamplesAppendix}
\end{figure*}

For our experiments in the main body, we have used a training set of 8800 synthetic images from the Cracktal simulator. Figure \ref{PlotsSetsize} illustrates the impact of training set size on the performance for the baseline U-Net model. Here we observe that after a set size of 4000 images the performance appears to stagnate on the $F1$ and $clDice$ metrics. On the $HDF_euc$, we observe an increase of almost $2$ between set size $4000$ and $8800$. It appears that our Cracktal simulator generates a good variety of examples and that a training set of around $8800$ is reasonable. 

As a complement, table \ref{cracktalLabel} further highlights the test set performance of different models on the in-distribution Cracktal data. Here we observe that the baseline U-Net slightly outperforms the other models on $F1$, $F1_{\theta=10}$, $HDF_{RBF}$. On the clDice and $HDF_{Euc}$ metrics, U-Net also achieves the best mean value performance,  but the obtained values lie within statistical deviation when contrasted with our CAP-Net design. As observed in the main body's results, the performance of U-Net drops more significantly on out-of-distribution data (multi-source set and SegCODEBRIM) compared to our proposed model CAP-Net. This shows that in-distribution performance is not predictive of the performance on out-of-distribution data. 

Furthermore, CAP-Net is more robust than the simple fully data-driven U-Net. Figure \ref{realExamplesAppendix} contains multiple qualitative examples of the baseline U-Net and CAP-Net. Here it is evident that our design choices improve the performance overall and help detecting cracks that U-Net misses.

\section{Limitations and Prospects}
In this section, we highlight potential remaining challenges and areas of improvement in our approach. \\

\noindent \textbf{Pointwise Mutual Information (PMI):} The computed PMI values depend on the chosen hyperparameters, such as the image scale or neighborhood size. While some solutions have been presented in the literature to address some of these challenges, such as scale invariance \cite{isola2014crisp}, these come with a substantial computational overhead. In contrast, the simplified PMI estimation presented in this work offers a good trade-off in terms of computational efficiency, especially when deployed in conjunction with neural networks. We also note that if the input image consists of extensive amount of cracked surface, where the cracked pattern itself forms a regular texture, PMI will dampen the response for crack detection (which no longer constitutes an anomaly). However, in real-world applications, we are interested in early onset detection of crack formation, before it's too late and structural collapse is imminent. Hence this situation is unlikely to be faced in real-world practice. \\

\noindent \textbf{Adaptive instance normalization (AdaIN):} We made use of textures from the Describable Textures dataset \cite{cimpoi14describing} to perform the style transfer. One immediate prospect for improvement lies in leveraging a concrete inspection dataset, which could potentially offer a more diverse range of textures that are even more relevant to the task at hand. \\

\noindent \textbf{The challenge of crack scale:}
Another possible challenge of our current system is the cracks' scale. If the cracks are too wide, our model has difficulty detecting it. This is mainly because we only generated thin cracks in Cracktal. Again, this focus stems from the fact that  inspectors are mainly interested in the early formation of cracks in real world scenarios. Wider cracks are usually found in concrete surfaces where the inspectors are already aware that the surface is damaged and were structures are severely damaged, potentially beyond repair. 

\newpage 
{\small
\bibliographystyle{ieee_fullname}
\bibliography{egbib}
}